\documentclass[final,3p,10pt]{elsarticle}
\makeatletter
\def\ps@pprintTitle{%
 \let\@oddhead\@empty
 \let\@evenhead\@empty
 \def\@oddfoot{\centerline{\thepage}}%
 \let\@evenfoot\@oddfoot}
\makeatother

\usepackage{amssymb}
\usepackage{lineno}
\usepackage{graphicx}
\usepackage{subfigure}
\usepackage{slashbox}
\usepackage{capt-of}
\usepackage{colortbl}
\usepackage[table]{xcolor}
\usepackage{url}

\begin{document}

\begin{frontmatter}

\title{Face Image Analysis using AAM, Gabor, LBP and WD features for Gender, Age, Expression and Ethnicity Classification}


\author{N. S. Lakshmiprabha}

\begin{abstract}
The growth in electronic transactions and human machine interactions rely on the information such as  gender, age, expression and ethnicity provided by the face image. In order to obtain these information, feature extraction plays a major role. In this paper, retrieval of age, gender, expression and race information from an individual face image is analysed using different feature extraction methods. The performance of four major feature extraction methods such as Active Appearance Model (AAM), Gabor wavelets, Local Binary Pattern (LBP) and Wavelet Decomposition (WD) are analyzed for gender recognition, age estimation, expression recognition and racial recognition in terms of accuracy (recognition rate), time for feature extraction, neural training and time to test an image. Each of this recognition system is compared with four feature extractors on same dataset (training and validation set) to get a better understanding in its performance. Experiments carried out on FG-NET, Cohn-Kanade, PAL face database shows that each method has its own merits and demerits. Thus it is practically impossible to define a method which is best at all circumstances with less computational complexity. Further, a detailed comparison of age estimation and age estimation using gender information is provided along with a solution to overcome aging effect in case of gender recognition. An attempt has been made in obtaining all (i.e. gender, age range, expression and ethnicity) information from a test image in a single go.
\end{abstract}


\end{frontmatter}

\setpagewiselinenumbers
\section{Introduction} 
Human Machine Interactions are increasing everyday, where a machine is taught to behave as human beings. Potential effort has been made in making a machine to perceive and infer information from a scene. A friendly environment is possible only by understanding end user's identity, mood, background (or ethnic group), gender, age group, body gesture etc. If a machine can change its attributes depending on the user's visual cues (ethnic, expression or age group) will attract more attention. This made many researchers to analyze face images for acquiring these information automatically. All these face processing techniques have potential application such as surveillance, human-machine interface, talking head, human emotion analysis, age synthesis, electronic customer relationship management, access control, marketing for example. The recent availability of relatively cheap computational power made face processing commercially available. 

Face is a complex 3D object and dynamic in nature. A face image encounters various problems such as pose, illumination and occlusion. The variation caused by these problem in a face image increases the difficulty in recognition process. Recognizing faces of their own race is more accurate than faces from other race \cite{raceeffect1}. In most cases other race face images looks very similar to each other. Training a system with particular racial background face images and then given a face image from other racial group lacks from recognizing the given face image correctly. This is because of other race effect and its effect on face recognition algorithm is studied in detail by Nicholas Furl et. al. \cite{raceeffect}. There are distinct variation in facial features depending on the country or background they belong \cite{race2}. It is easy to differentiate an american from an asian. This is mainly due to the factors such as weather condition, food habits, life style, hereditary etc. Racial features in face images also varies between individuals of different age group and gender \cite{race3}. There are many methods which work well for particular background people wherein suffers with other group. In particular, color based face detection method fails to detect black originated people, since it is influenced by the color range. A significant application would be, a machine with a capability to change its communicating language to the most familiar language spoken in a particular country using racial or ethnic information from a person's visual cues.

Face also varies to a large extend with different expressions, since face is the index of mind. There are six primary emotions which are distinct from each other and these emotions are also called as basic emotions. These basic emotions include happy, sad, fear, disgust, angry, and surprise \cite{exp1}. The facial expression recognition is broadly classified into Image-based, Model-based, Motion extraction \cite{exp2}. A comprehensive survey of the currently available databases that can be used in facial expression recognition systems is presented in \cite{exp3}. Most commonly used databases include Cohn-Kanade facial expression database, Japanese Female Facial Expression (JAFFE) database, MMI database and CMU-PIE database. The way a human begin express his/her emotion varies considerably in male and female, different ethnic group and different age group. 

\begin{figure}[hbt]
\begin{center}
\subfigure[Male and Female face from FG-NET \cite{fgnet}.] {\label{malefemale}\includegraphics[scale=0.4]{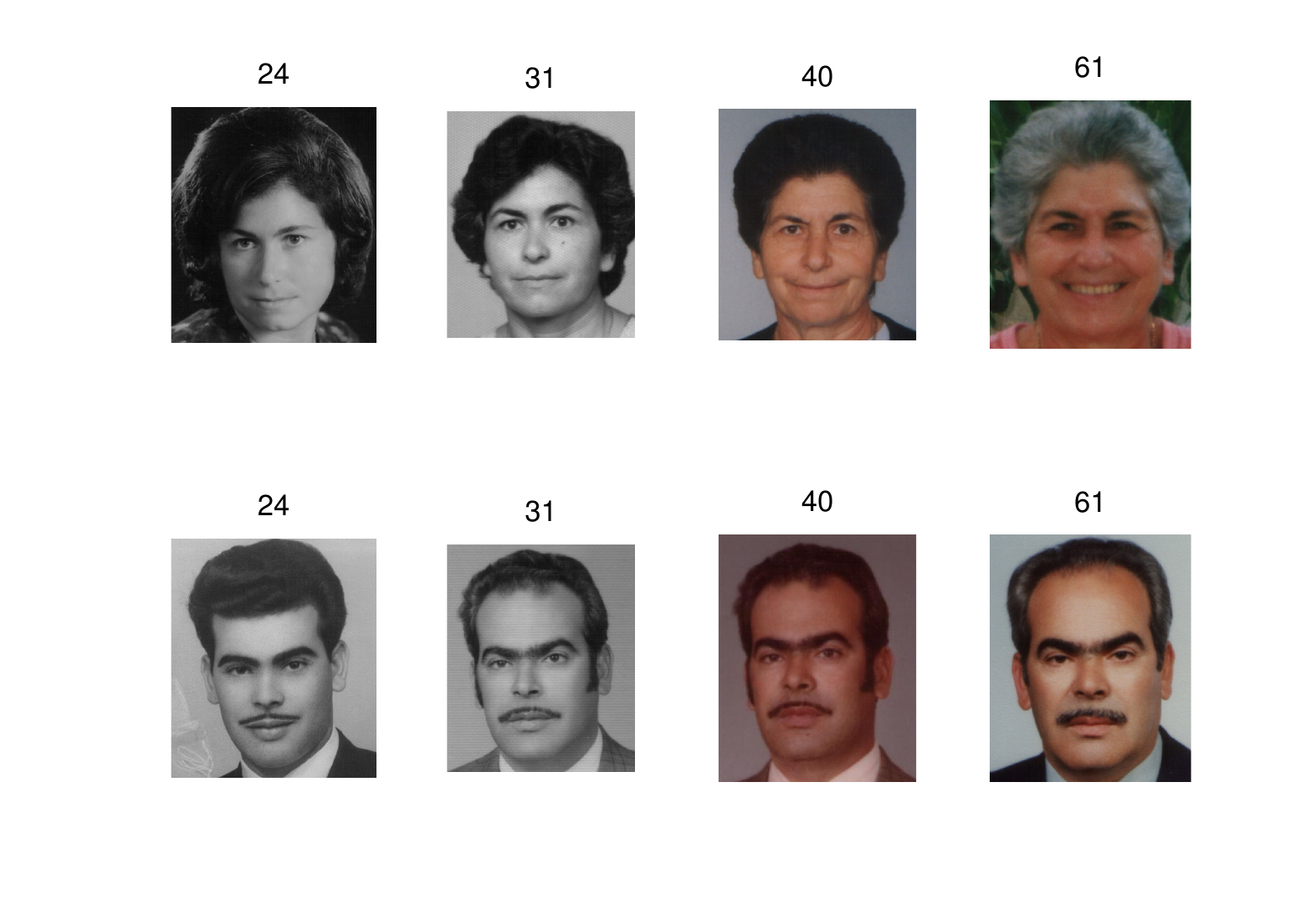}}
\subfigure[Block Diagram.] {\label{block1}\includegraphics[scale=0.35]{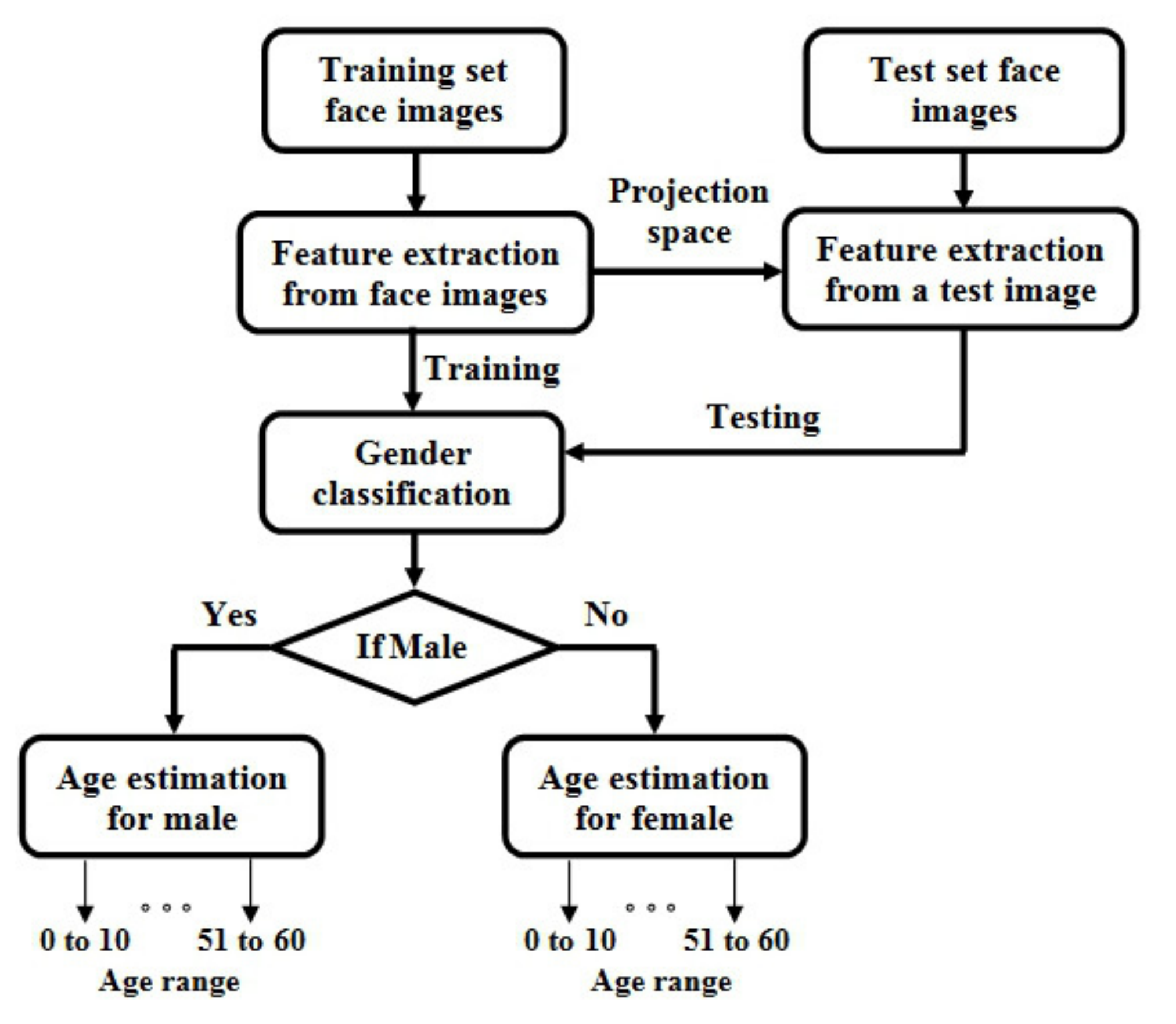}}
\end{center}
\caption{Age Estimation using Gender Information.}
\label{block11}
\end{figure}
The human faces are subjected to growth or aging change which is one of the most non-linear changes occurring on the face. This change in appearance vary from person to person and it is contributed by various factors like ancestry, health, lifestyle, race, gender, working environment, climate, decrease or increase in weight, emotional stress, smoking, drug use, diet, and emotional stress \cite{ageest1,ageest3}. Males and Females may age differently as they have varying type of face aging pattern \cite{ageest1,ageest2}. This is due to the difference in makeup, hair style, accessories in female or mustache and beard in case of males. In adulthood, female faces appear younger than male faces. Figure \ref{malefemale} shows the face images of male and female with age labeled at the top. In order to overcome this, Gender classifier and age estimation blocks are cascaded as shown in figure \ref{block1} \cite{mypaper}. Depending upon the output of the gender classifier, the appearance parameter is fed to male or female age estimator. Further, recognition rate of gender classifier decreases if there are aging variation in the face images \cite{ageeffect}. 

This paper deals with four feature extraction methods namely Active Appearance Models (AAM) \cite{AAM1}, Gabor wavelets \cite{ref25}, Local Binary Pattern (LBP) \cite{exp5} and Wavelet Decomposition (WD, also called as Discrete Wavelet Transform) \cite{WDref1,WDref2,wdref4} for gender recognition, age estimation, expression recognition and racial recognition. Neural network is used as classifier throughout this paper. The analysis is made in terms of accuracy and time consumption. Given a face image, obtaining gender, age, expression and ethnicity in a single go within a second to compute all those information is of particular interest. There are research work which illustrated the effectiveness of a single method on expression, face, gender recognition and age estimation \cite{all1,all2,all3}. But Not much work has been done in retrieving all these information from a test image. This has got potential application such as surveillance, marketing (for accessing which particular product is preferred by which age group), proventing access to web sites and refining the database search etc. In this work, above mentioned four features are examined in obtaining all these information from a test image. Face recognition (i.e. identity) is not included because there is no database which provide all the above said information along with more than two images per person (at-least one image for training and the other for testing).

\begin{figure}[htb]
\centering
\includegraphics[width=8cm]{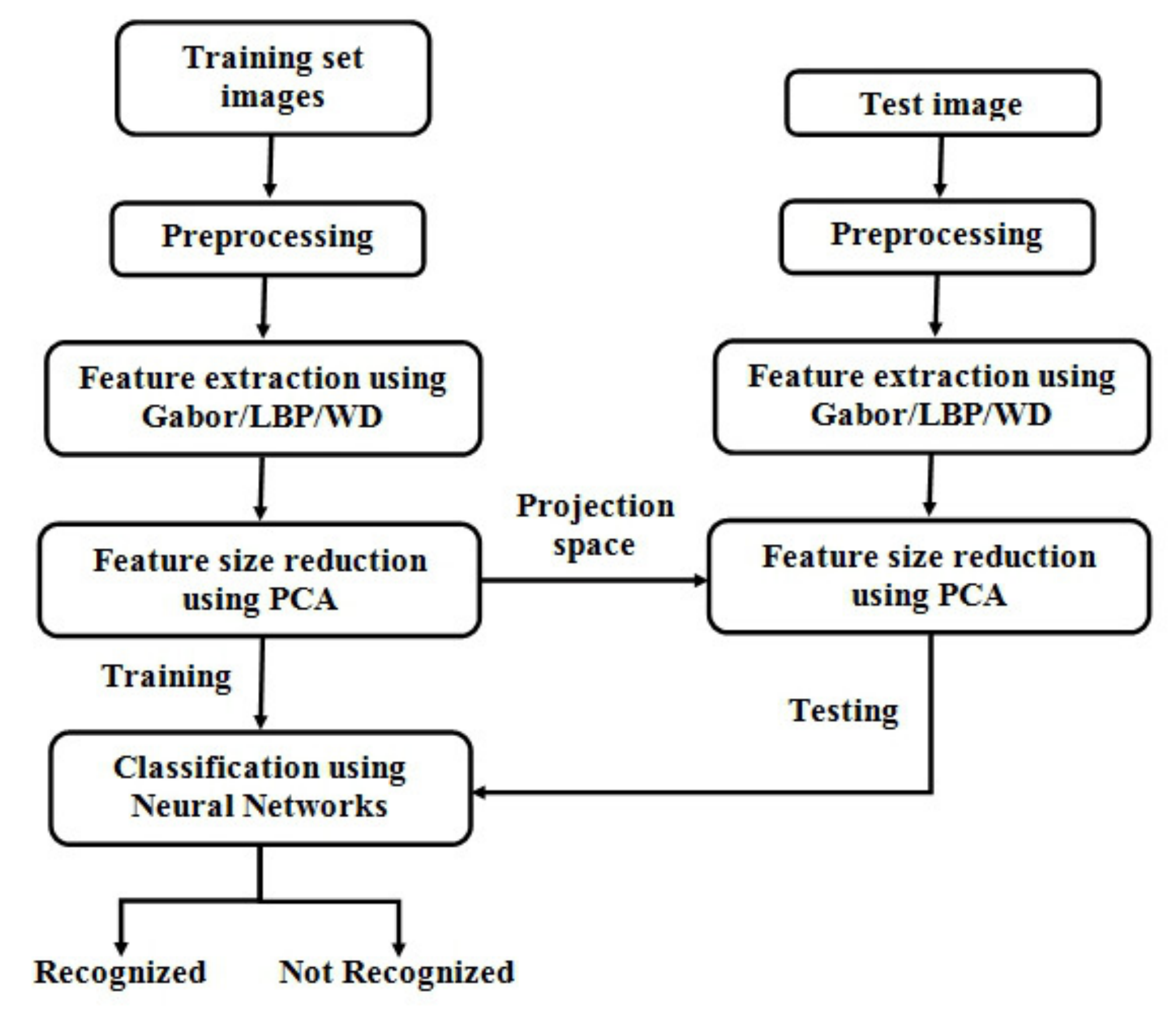}
\caption{Block diagram of Gabor/LBP/WFB-PCA-Neural Network method.}
\label{block2}
\end{figure}

Each of this AAM, Gabor, LBP and WD feature extractor has been applied for gender recognition \cite{genexp,gengab,ageeffect,genlbp}, age estimation \cite{refage10,ref11,agelbp,agewd}, racial recognition \cite{race2,racelbp} and expression recognition \cite{AAMexp1,jaffe,gabexp2,exp7,exp8,expwd,expwd2}. AAM extracted appearance parameters are good in providing global shape and texture information. Whereas Gabor wavelet, LBP and WD are rich in local shape and texture features. These three methods offer rich features with larger feature vector size. By using regular spacing grids, the size of the feature vectors can be reduced. In many cases there will be a demand to consider more features. Size of the feature vector is increased by reducing the grid spacing also results in increased time consumption, complexity and storage requirement. As well these local features are prone to noise. By performing Principal Component Analysis (PCA) \cite{ref10} on the feature vectors these issues can be addressed. The resulting feature vector size will be less than or equal to the number of images in database. The recognition rate will also increase because the problem of PCA under illumination variation is eliminated by these features. The problem with PCA is, it finds variation irrespective of class membership which has been solved by combining Neural networks as classifier. Simple block diagram of Gabor/LBP/WD-PCA-Neural Network method is shown in figure \ref{block2}.

This paper is organized in the following way, Section \ref{preprocess} explains about the preprocessing and normalization. Section \ref{feature} elaborates on AAM, Gabor, LBP and WD feature extraction methods. Section \ref{pca} elucidates on the feature dimension reduction step required for Gabor/LBP/WD features. Classification using neural networks is explained in section \ref{neural}. Section \ref{result} discusses on the experimental results.

\section{Preprocessing and Normalization}
\label{preprocess}
\begin{figure}[htp]
\begin{center}
\subfigure[Database face images.] {\label{norm1}\includegraphics[scale=0.46]{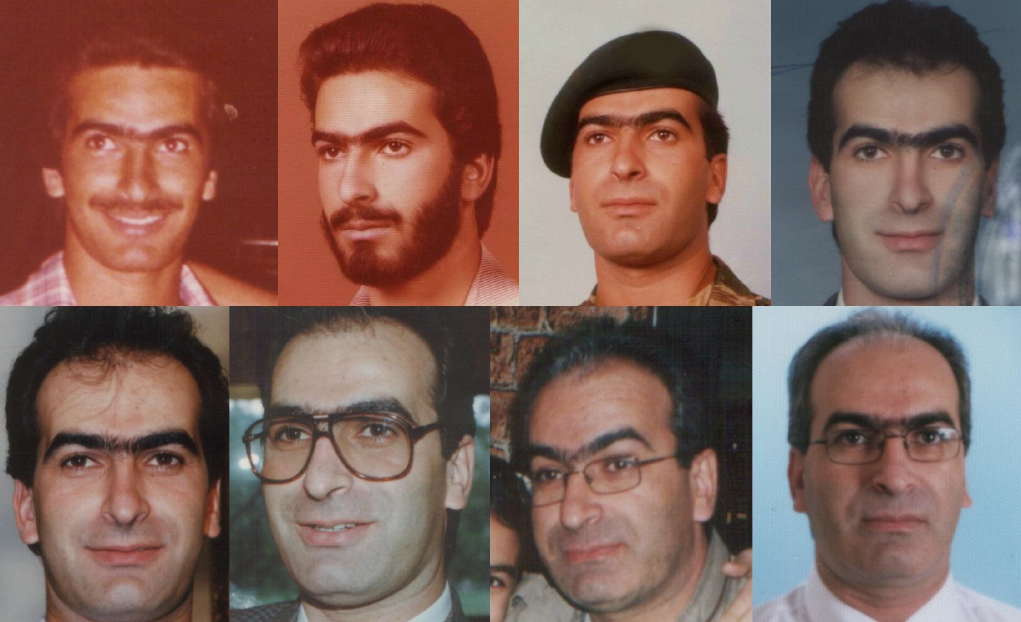}}
\subfigure[Normalized face images.] {\label{norm2}\includegraphics[scale=0.8]{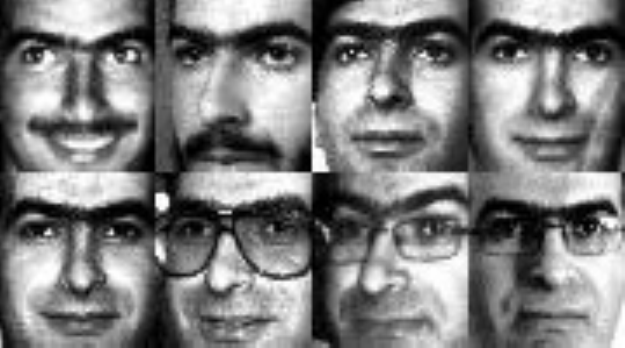}}
\end{center}
\caption{Face images from FG-NET database.}
\label{norm}
\end{figure}

The size of the images in the database is larger and contains background information in many cases. This information is irrelavant and to avoid this image normalization is performed. Pixel location of eye center is used for face normalization followed by histogram equalization. The images from FG-NET database \cite{fgnet} and the normalized face images (size 65x60 pixels) are shown in figure \ref{norm}. Preprocessing step is required for Gabor, LBP and WD feature extraction methods.

\section{Feature Extraction Methods}
\label{feature}
The four feature extraction methods namely Active Appearance Models (AAM) \cite{AAM1}, Gabor wavelets \cite{ref25}, Local Binary Pattern (LBP) \cite{exp5} and Wavelet Decomposition (WD) \cite{WDref1,WDref2,wdref3,wdref4} is discussed in this section. 

\subsection{Active Appearance Model (AAM)}
\label{secAAM}
\begin{figure}[htp]
\begin{center}
\begin{minipage}{0.35\textwidth}
\includegraphics[width=\textwidth]{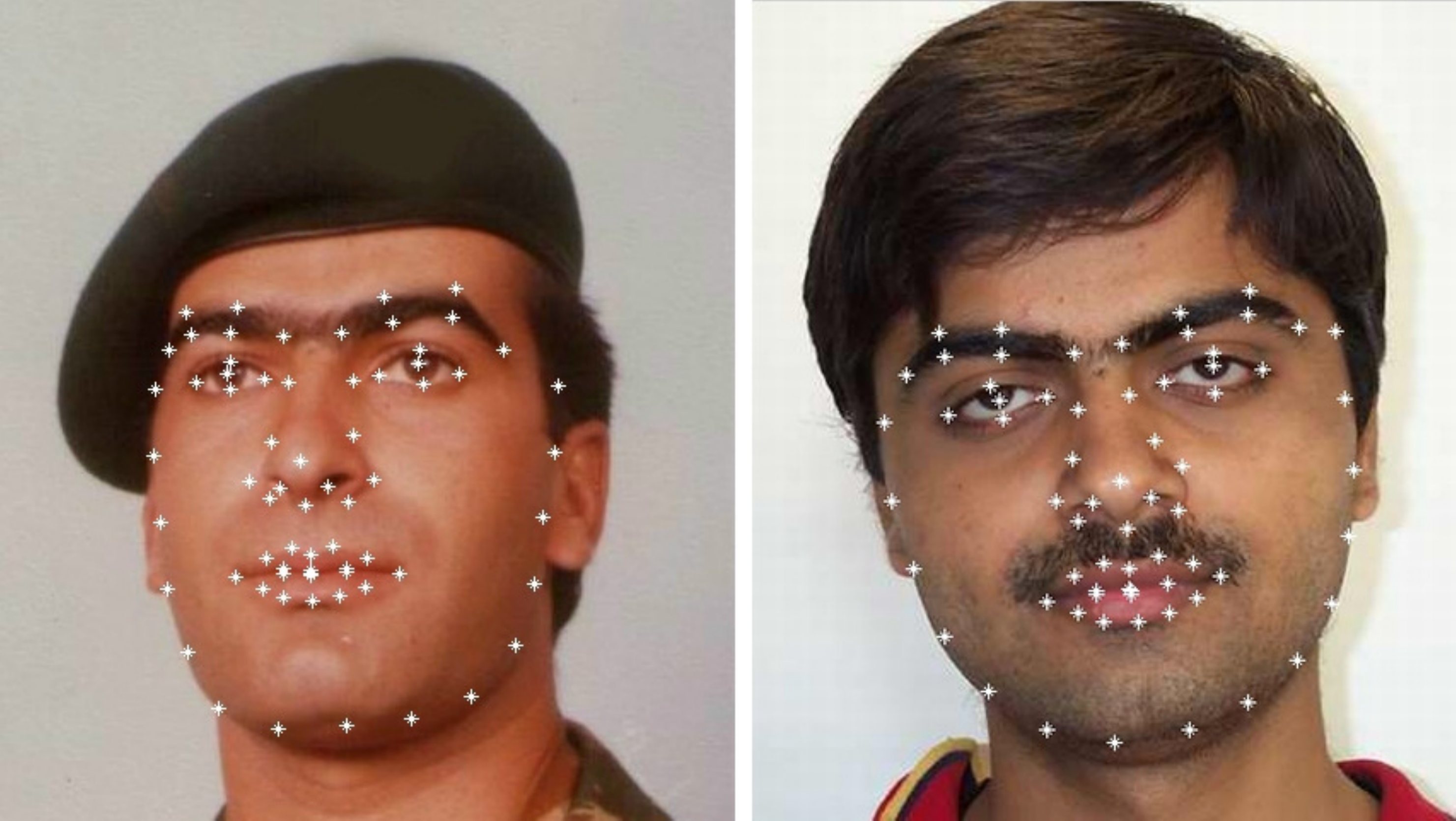}
\captionof{figure}{Hand annotated face images from FG-NET and PAL database.}
\label{fig3}
\end{minipage}
\hspace{0.5cm}
\begin{minipage}{0.45\textwidth}
\centering
\begin{tabular}{|c|c|}
\hline
\rowcolor[gray]{.8}
Landmark points & Location\\
\hline
1-15 & Face outer\\
\hline
16-21 & Right eyebrow\\
\hline
22-27 & Left eyebrow\\
\hline
28-32 & Left eye\\
\hline
33-37 & Right eye\\
\hline
38-48,68 & Nose\\
\hline
49-67 & Lips\\
\hline
\end{tabular}
\captionof{table}{Landmark point detail.}
\label{fig3tab}
\end{minipage}
\end{center}
\begin{center}
\begin{minipage}{0.35\textwidth}
\includegraphics[width=\textwidth]{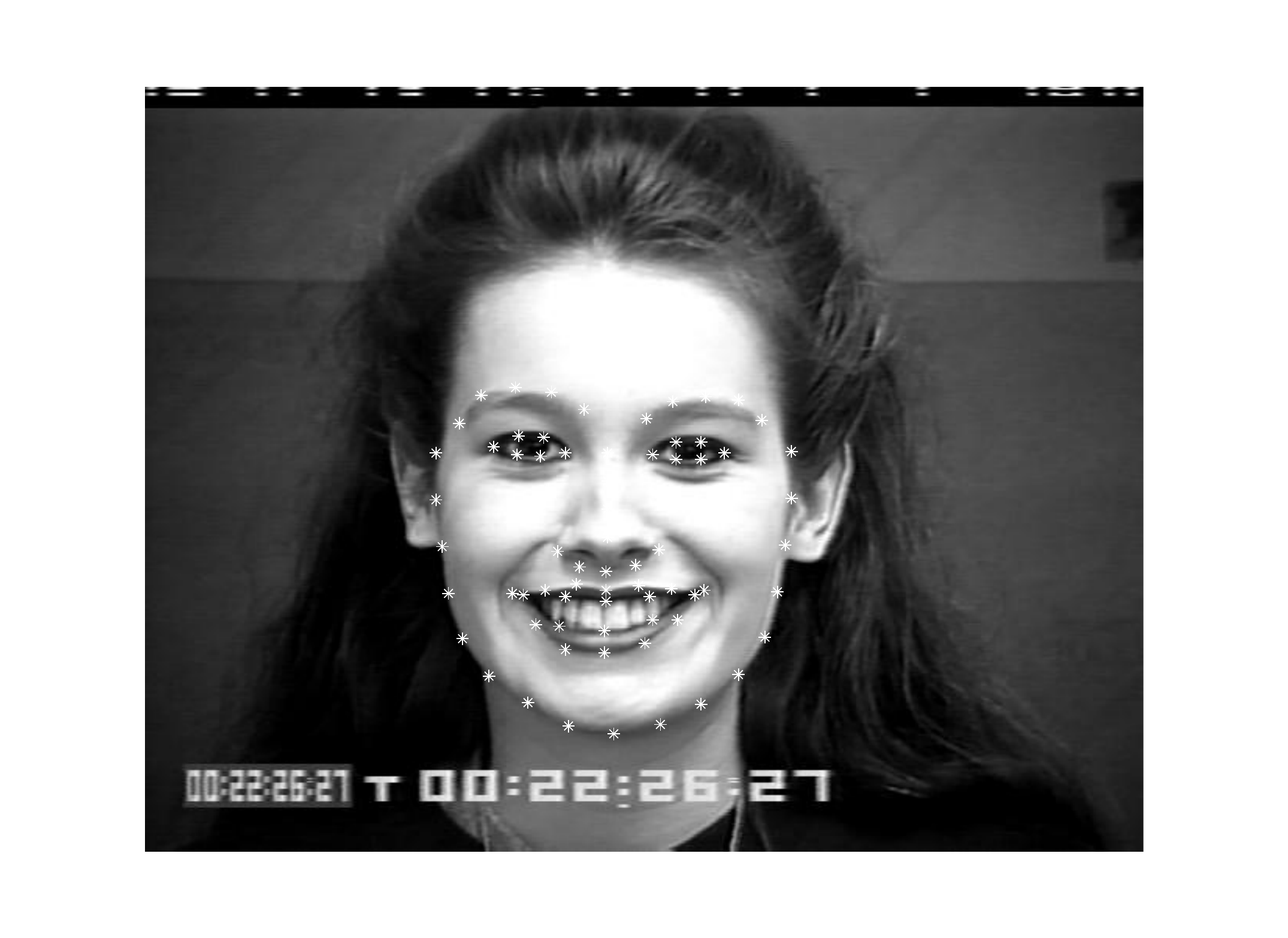}
\captionof{figure}{Hand annotated face images from Cohn-Kanade.}
\label{fig9a}
\end{minipage}
\hspace{0.5 cm}
\begin{minipage}{0.45\textwidth}
\centering
\begin{tabular}{|c|c|}
\hline
\rowcolor[gray]{.8}
Landmark points & Location\\
\hline
1-17 & Face outer\\
\hline
18-22 & left eyebrow\\
\hline
22-26 & Right eyebrow\\
\hline
27-36 & Nose\\
\hline
37-42 & left eye\\
\hline
43-48 & Right eye\\
\hline
49-68 & Lips\\
\hline
\end{tabular}
\captionof{table}{Landmark point detail.}
\label{fig9atab}
\end{minipage}
\end{center}
\end{figure}

Active Appearance Model \cite{AAM1,ASM1} is a statistical model of appearance. The feature extracted from AAM has both shape and texture information which is more suitable for aging variations. The training set consists of hand annotated face images. The hand annotated face images of FG-NET, PAL and Cohn-Kanade database is shown in figure \ref{fig3}, \ref{fig9a} and landmark point details are given in table \ref{fig3tab}, \ref{fig9atab} respectively. Different shape landmark points also influences the performance and its effects are discussed in results section (see section \ref{result}).

\begin{figure}[htp]
\begin{center}
\subfigure[FG-NET Database.] {\label{fig5}\includegraphics[scale=0.43]{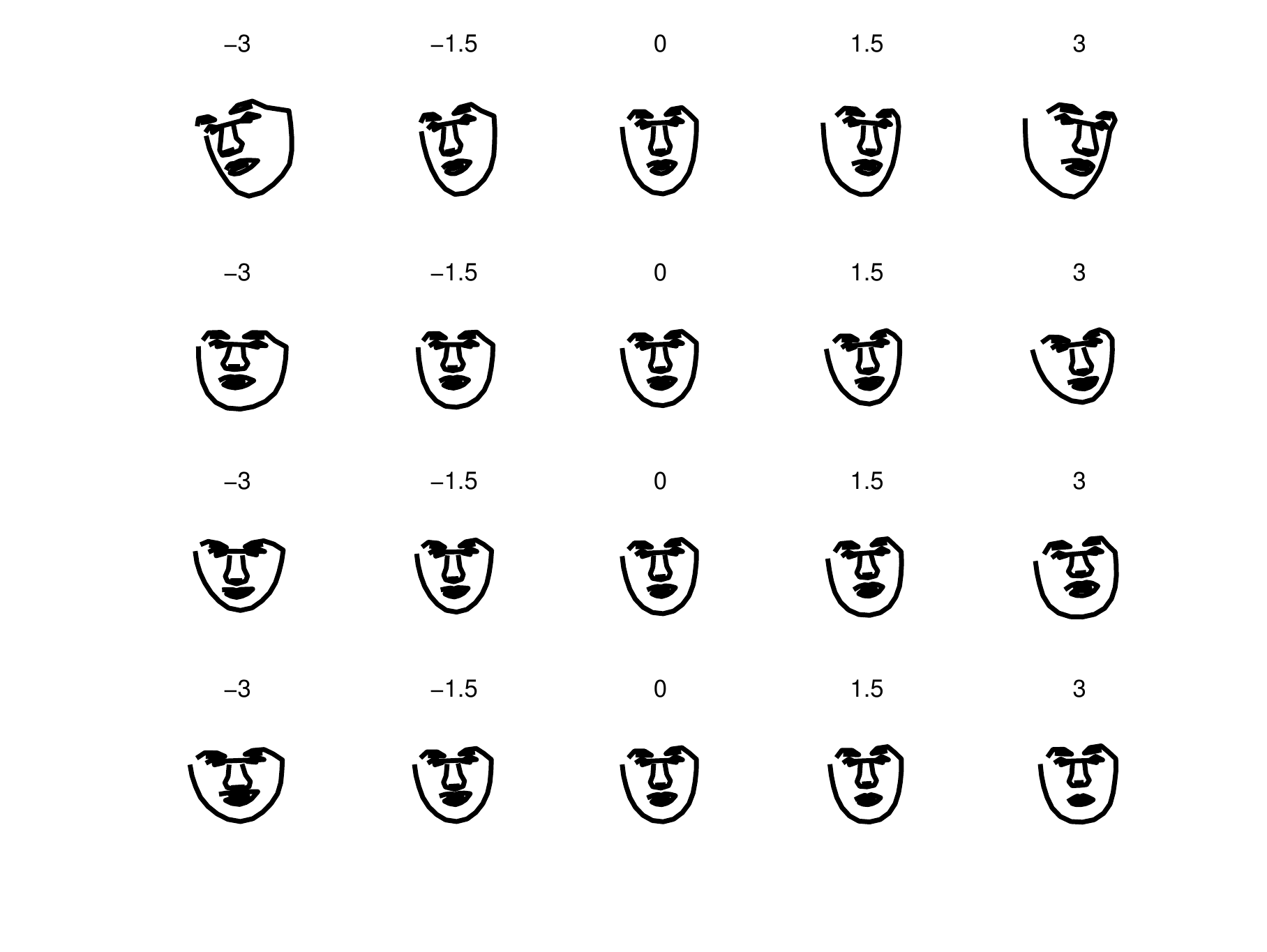}}
\subfigure[Cohn-Kanade Database.] {\label{fig6}\includegraphics[scale=0.43]{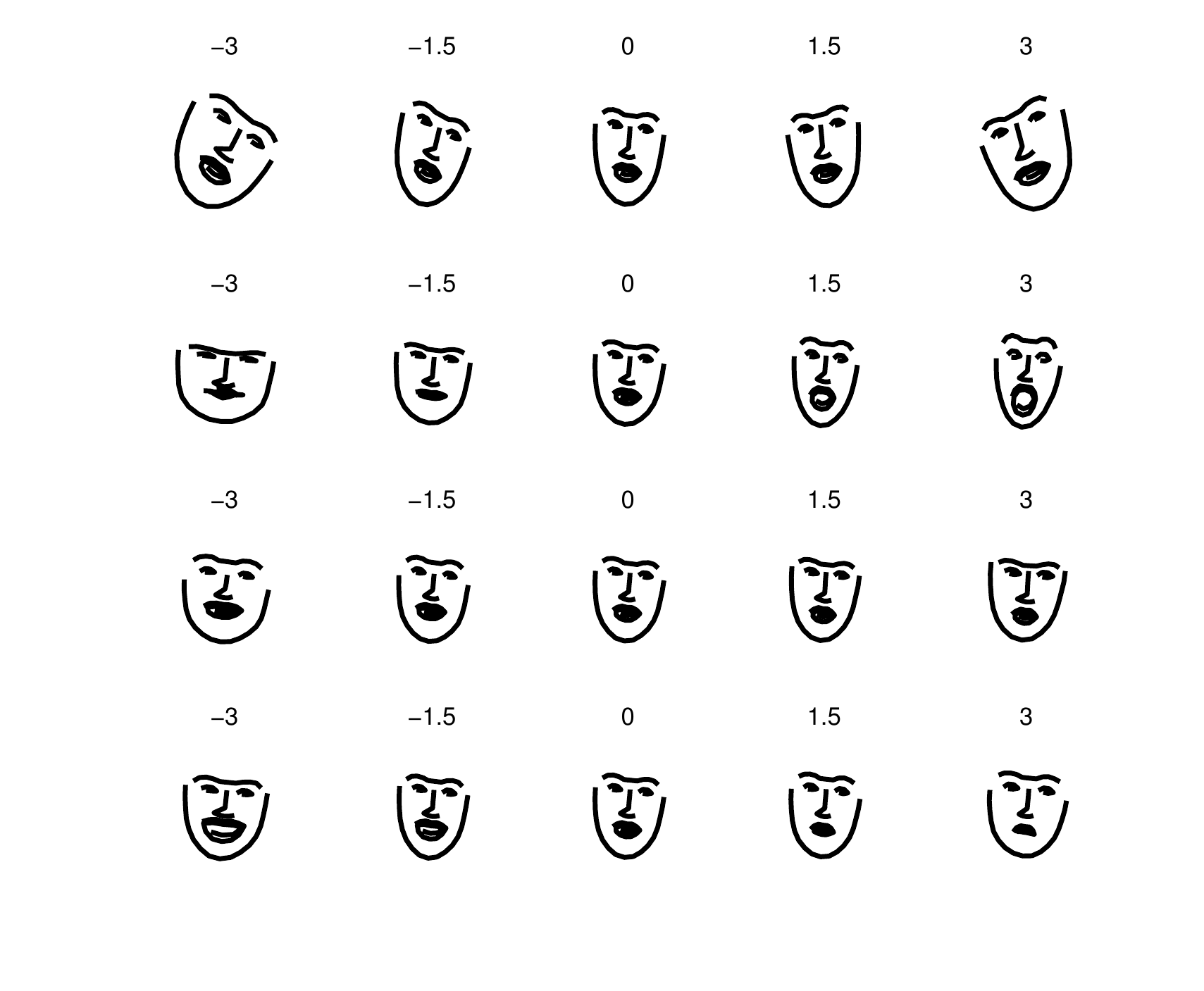}}
\end{center}
\caption{Shape changes obtained by varying mean shape.}
\label{shape}
\end{figure}

Let $I = [I_1, I_2, \dots , I_N]$ represents N training set images with landmark points as $x = [x_1, x_2, \dot , x_N]$. Shape variations are obtained by aligning these landmark points and then Principal Components Analysis (PCA) is performed on those points. Any shape vector x in the training set can be represented as in equation (\ref{equ1}).
\begin{equation}
\label{equ1}
x \approx \overline{x} + V_sb_s
\end{equation}
where $\overline{x}$ is the mean shape, $V_s$ contains the eigenvectors of largest eigenvalues ($\lambda_s$) and $b_s$ represents weights or shape model parameters. By rewriting equation (\ref{equ1}), it is possible to calculate shape model parameters corresponding to a given example. The shape can be changed by varying the elements of $b_s$ using eigenvalues ($\lambda_s$). Figure \ref{shape} shows the shape changes obtained by applying limits of $\pm 3\sqrt{\lambda_s}$ to the mean shape. Center face shape in figure \ref{shape} indicates the mean shape ($\overline{x}$) of all the training set images.
\begin{equation}
\label{equ2}
b_s = V_s^T(x - \overline{x})
\end{equation}

\begin{figure}[htb]
\centering
\includegraphics[width=6cm]{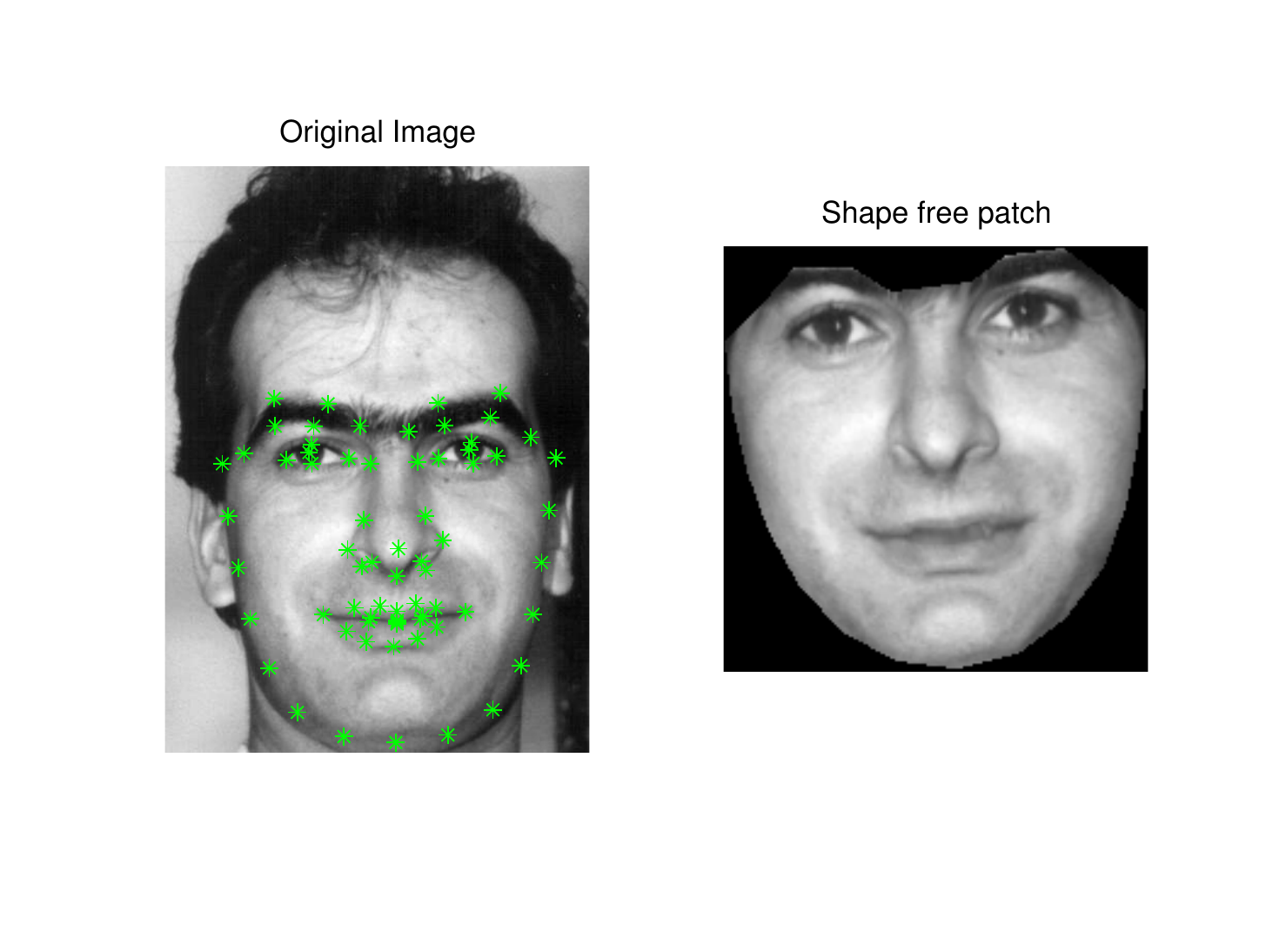}
\caption{An example shape free patch.}
\label{shapefree}
\end{figure}

Shape free patch is obtained by warping each training set images into the mean shape. Size of the warped texture also place a major role and it is discussed in experimental results section \ref{result}. An example shape free patch is shown in figure \ref{shapefree}. 

Let $g = [g_1, g_2, \dots , g_N]$ be the shape free vectors of all the training set images. Similar to shape modeling, texture modeling is done by using PCA.
\begin{equation}
\label{equ3}
b_g = V_g^T(g - \overline{g})
\end{equation}
where $b_g$ is the weights or grey-level model parameter, $V_g$ is the eigenvectors and $\overline{g}$ is the mean grey-level vector.

\begin{figure}[htp]
\begin{center}
\subfigure[FG-NET Database.] {\label{fig7}\includegraphics[scale=0.5]{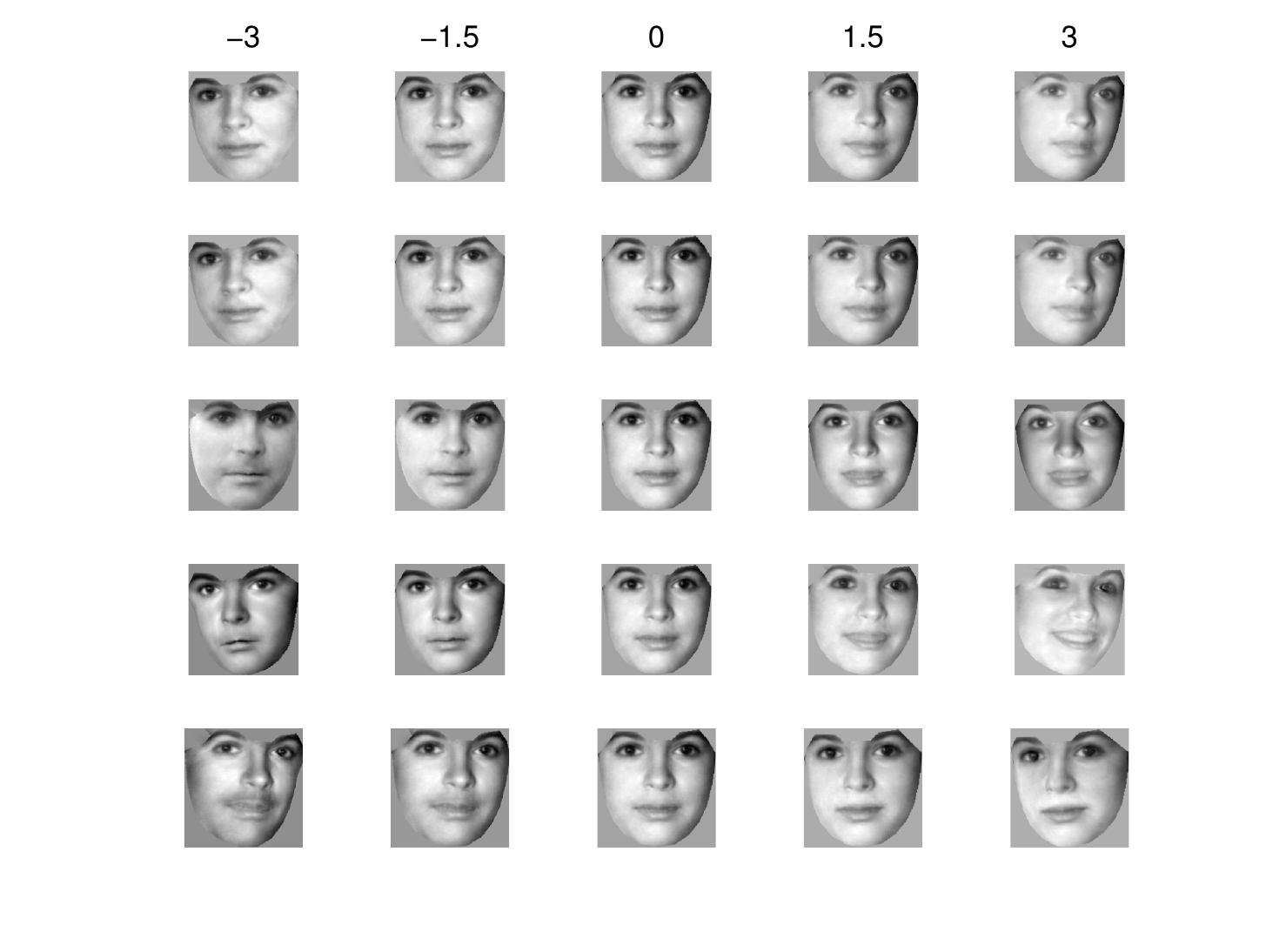}}
\subfigure[Cohn Kanade Database.] {\label{fig8}\includegraphics[scale=0.5]{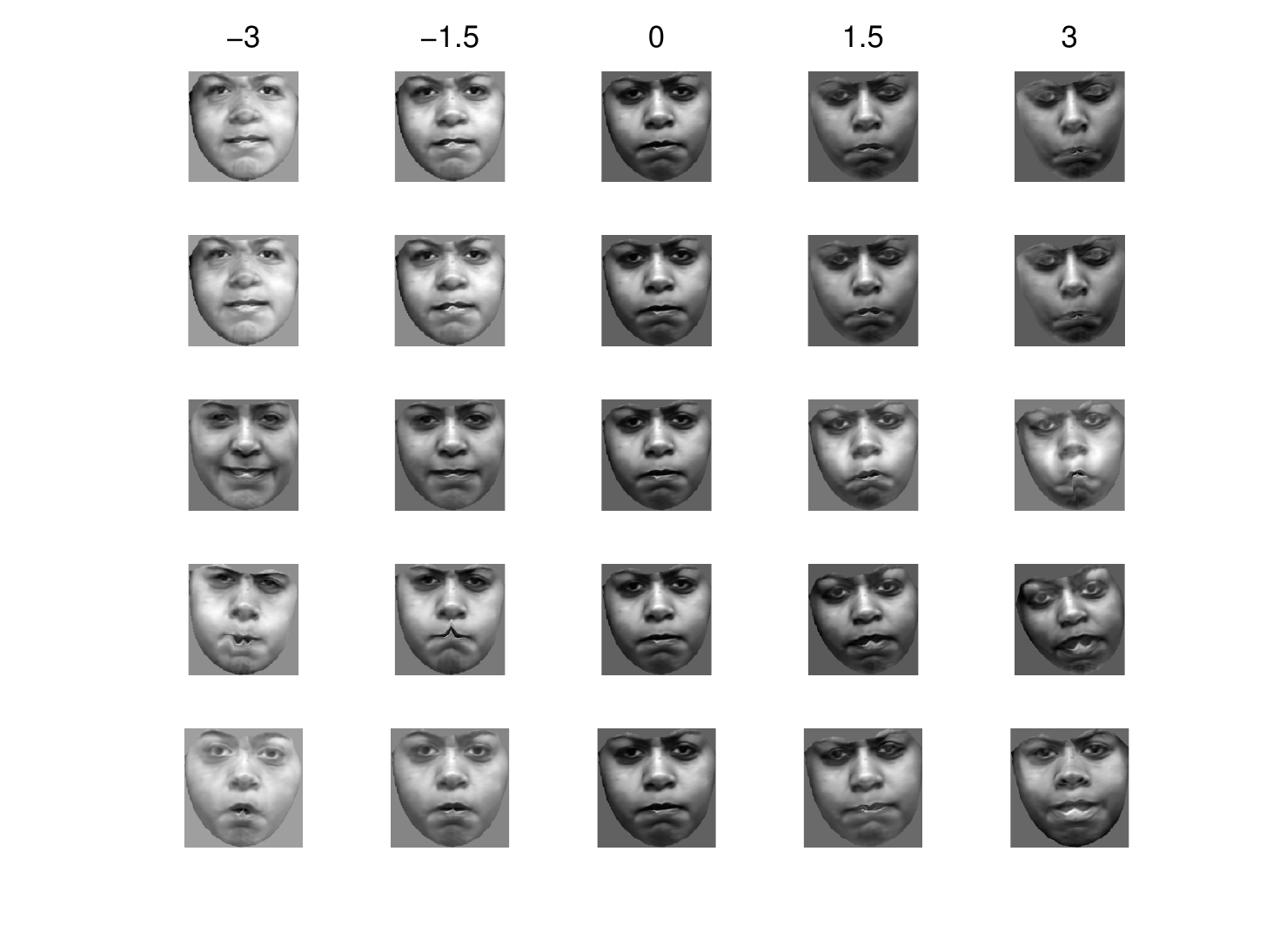}}
\end{center}
\caption{Effect of varying first five appearance parameters.}
\label{appearance}
\end{figure}
Appearance model parameter can be obtained by combining shape model parameter and grey-level model parameter. Since $b_s$ has distance as unit and $b_g$ has intensity as unit they cannot be compared together. $b_s$ and $b_g$ are made commensurate by using $W_s$. Combined parameter vector is obtained by using equation (\ref{equ4}). $W_s$ is a diagonal matrix of weights for each shape parameter. PCA is applied on combined parameter vector and the appearance parameter controlling both shape and texture of the model is calculated.
\begin{equation}
\label{equ4}
b_{sg} = \left(
\begin{array}{c}
W_sb_s\\
b_g
\end{array}
\right) 
\end{equation}
\begin{equation}
\label{equ5}
b_{sg} = Qc
\end{equation}
where $Q= \left(
\begin{array}{c}
Q_s\\
Q_g
\end{array}\right)$ and c is the appearance parameter. By varying c, it is possible to achieve changes in both shape and texture. Figure \ref{appearance} shows the effect of varying first five appearance parameters by $\pm 3\sqrt{\lambda_{sg}}$, where $\lambda_{sg}$ is the eigenvalues. As it is seen from figure \ref{fig5} and \ref{fig7}, there are huge variations in pose, head angle, expression and illumination. But in case of Cohn-Kanade database, figure \ref{fig6} and \ref{fig8} has variation majorly in expression. A point to be observed here is training data plays a major role and more variation in training images results in better appearance parameters.

When an annotated test image $(x_{test})$ is given as input, it is converted into shape model parameter $b_{stest}$ using equation \ref{equ2} and multiplied with $W_s$. The test image is warped with the mean shape and converted into shape free patch. Using equation \ref{equ3}, grey-level model parameter $b_{gtest}$ is calculated. Combining $b_{stest}$ and $b_{gtest}$ results in $b_{sgtest}$ and the appearance parameter $c_{test}$ is obtained using equation \ref{equ5}. $c_{test}$ is used for classification purpose.

\subsection{Gabor wavelet}
\label{gabor}
Local features in face images are more robust against distortions such as pose, illuminations etc. Spatial-frequency analysis is often desirable to extract such features. With good characteristics of space-frequency localization, Gabor wavelet is a suitable choice for recognition purpose. The Gabor wavelets (kernels, filters) \cite{ref25} can be defined as follows: 
\begin{equation}
\label{gabor1}
\psi_{\mu, \nu}(z) = \frac{||k_{\mu, \nu}||^2}{\sigma^2} e^{- \frac{||k_{\mu, \nu}||^2 ||z||^2}{2\sigma^2}} [e^{ik_{\mu, \nu}z} - e^{-\frac{\sigma^2}{2}}]
\end{equation}
Where $\mu$ and $\nu$ define the orientation and scale of the Gabor kernels,  the wave vector $k_{\mu, \nu}$, is defined as follows:
\begin{equation}
\label{gabor2}
k_{\mu, \nu} = k_{\nu}e^{i\phi_\mu}
\end{equation}
where $k_\nu= k_{max}/f^\nu$ and $\phi_\mu= \pi\mu/8$. $k_{max}$ is the maximum frequency, f is the spacing factor between kernels in the frequency domain and z =(x, y), $||\cdot||$ denotes the norm operator. 
Gabor wavelets at five different scales, $\nu[0,…,4]$, and eight orientations, $\mu[0,…,7]$ are considered in this work with the following parameters: $\sigma = 2\pi$, $k_{max} = \pi/2$ and $f = \sqrt{2}$. The size of all this filters are 32 x 32 pixels.

\begin{figure}[htp]
\begin{center}
\subfigure[Real part of the convolved face image.] {\label{fig1}\includegraphics[scale=0.45]{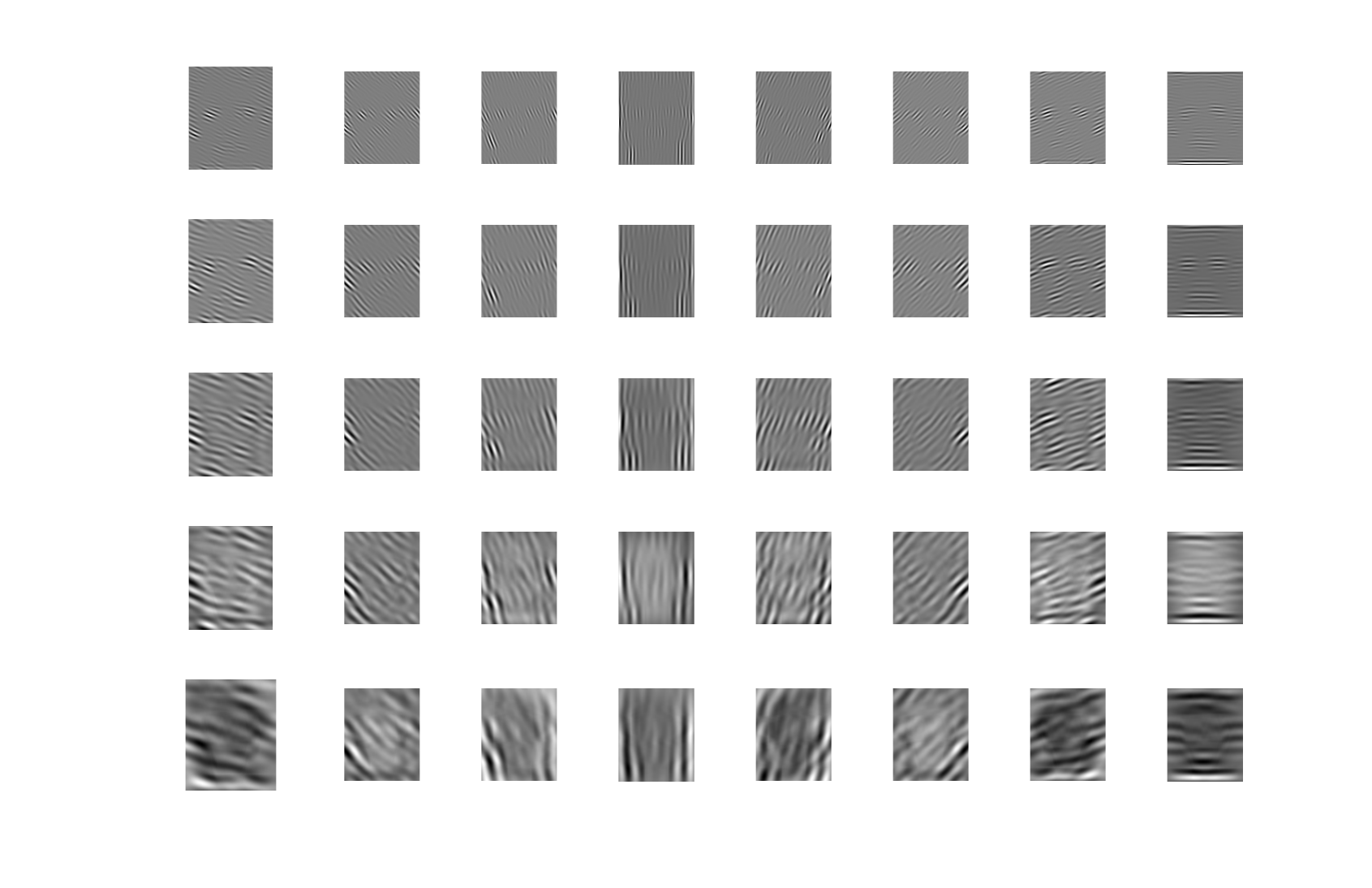}}
\subfigure[Magnitude of the convolved face image.] {\label{fig2}\includegraphics[scale=0.45]{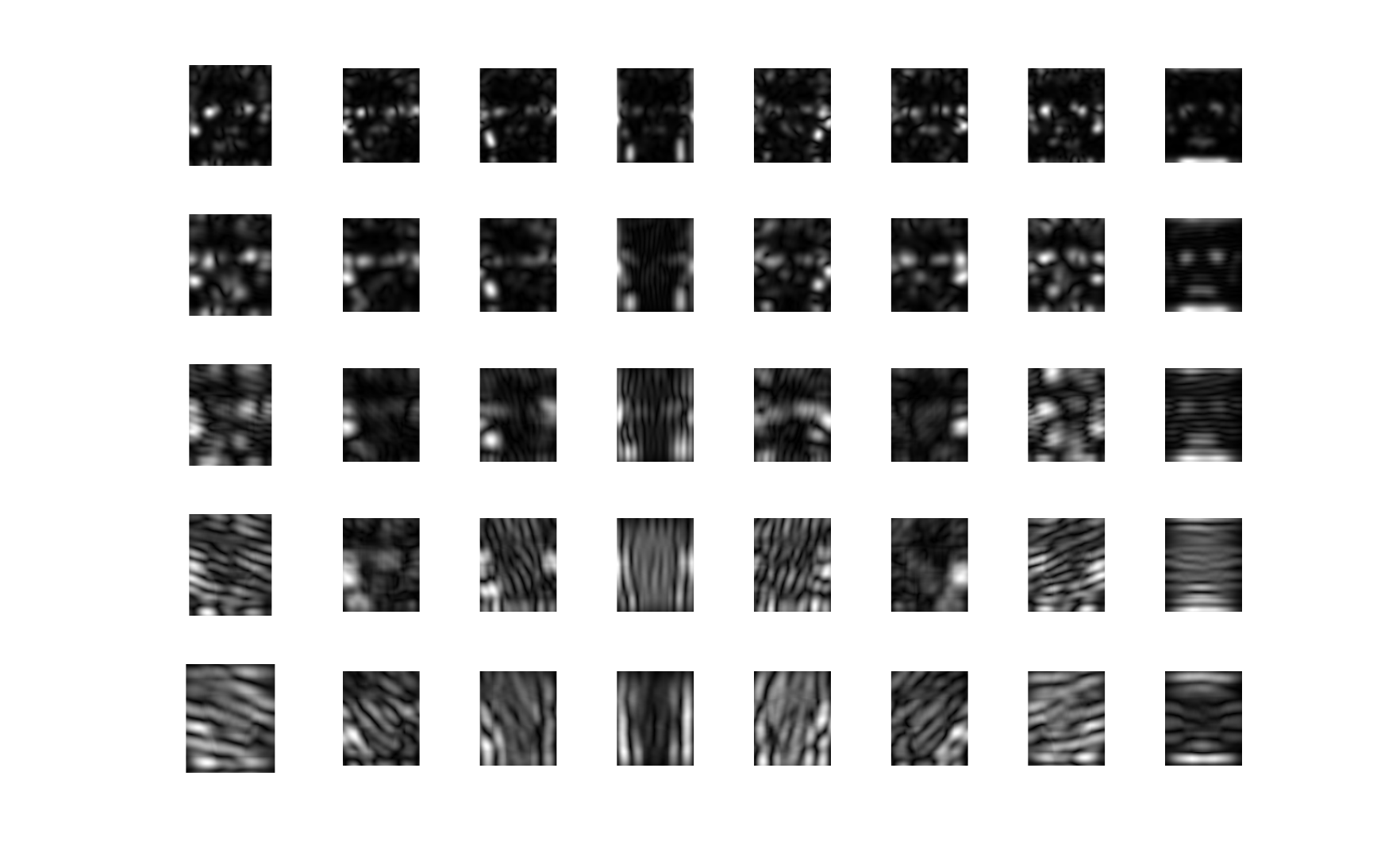}}
\end{center}
\caption{Real part and magnitude of the convolved face image.}
\label{edge}
\end{figure}
\begin{figure}[htb]
\centering
\includegraphics[width=8cm]{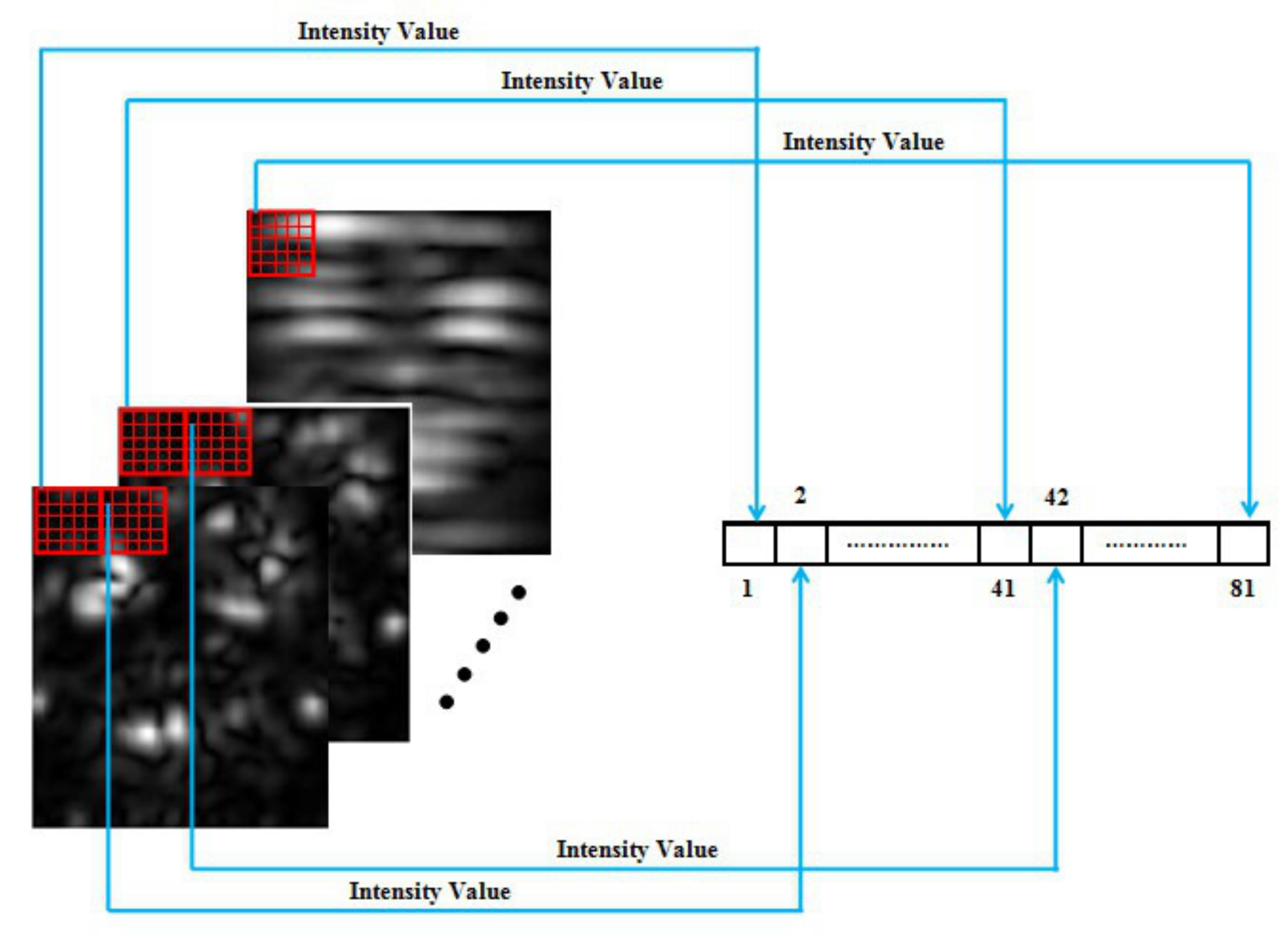}
\caption{Sampling feature vector from the 40 gabor convolved face images.}
\label{gabor3}
\end{figure}

An image can be represented in wavelet response by convolving Gabor filters of different scale and orientation. The set of convolution coefficients for kernels at one image pixel is called a jet. The resulting output contains most important face features like eyes, mouth and nose edges, as well as moles, dimples and scars. Real part and magnitude of convolved face image from ORL database is shown in figure \ref{fig1} and \ref{fig2} respectively. For each image after convolution there are 40 images containing extracted features. All these 40 images are converted into a feature vector. This also increases the time consumption and memory requirements. Huge feature vector size can be avoided by taking limited number of pixels from the feature images with regular spacing grids as shown in figure \ref{gabor3}.

\subsection{Local Binary Pattern (LBP)}
\label{LBP}
Local Binary Patterns provides a powerful means of texture description \cite{exp5}. LBP features are gray scale and rotation invariant texture operator. These features are more widely used for expression recognition \cite{exp7,exp8}. LBP feature extraction is faster than Gabor wavelet method and also provides similar performance.

\begin{figure}[htb]
\centering
\includegraphics[width=5in]{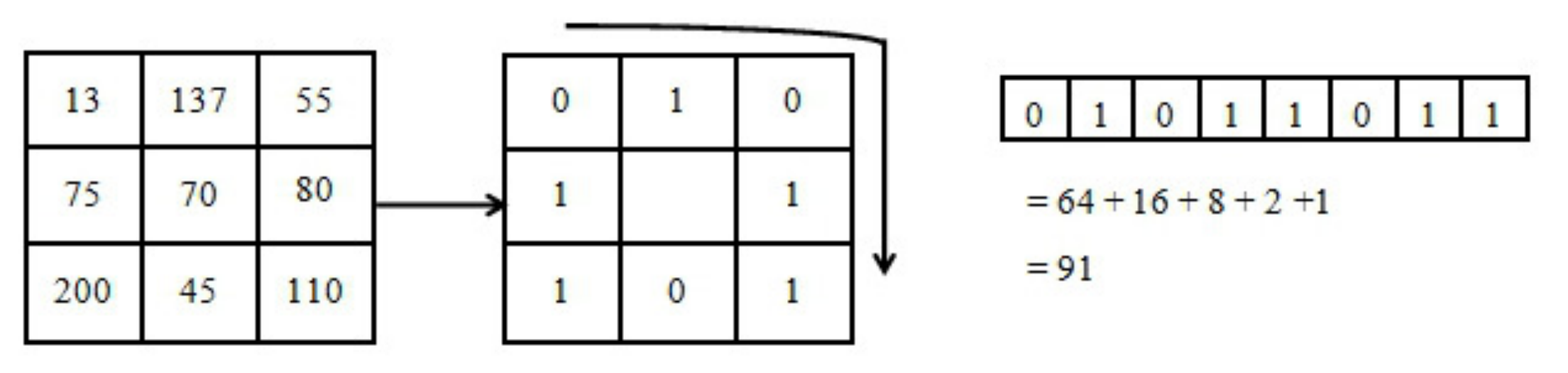}
\caption{Feature extraction using LBP.}
\label{chap5-fig9}
\end{figure}

Consider a 3x3 pixels with center pixel $(x_c, y_c)$ intensity value be $g_c$ and local texture as $T = t(g_0, \dots, g_7)$ where $g_i(i = 0, \dots, 7)$ corresponds to the grey values of the 8 surrounding pixels. These surrounding pixels are thresholded with the center value $g_c$ as $t(s(g_0 - g_c), \dots , s(g_7 - g_c))$ and the function s(x) is defined as,
\begin{equation}
s(x) = \left\{
\begin{array}{lr}
1 & , x > 0\\
0 & , x \le 0
\end{array}
\right.
\end{equation} 
Then the LBP pattern at a given pixel can be obtained using equation (\ref{chap5-lbp2}). An example of LBP operator is shown in figure \ref{chap5-fig9}.
\begin{equation}
LBP(x_c, y_c) = \sum_{i=0}^{7} s(g_i - g_c) 2^i
\label{chap5-lbp2}
\end{equation}

\begin{figure}[htp]
\begin{center}
\subfigure[LBP histogram features for a face image.] {\label{chap5-fig10}\includegraphics[scale=0.35]{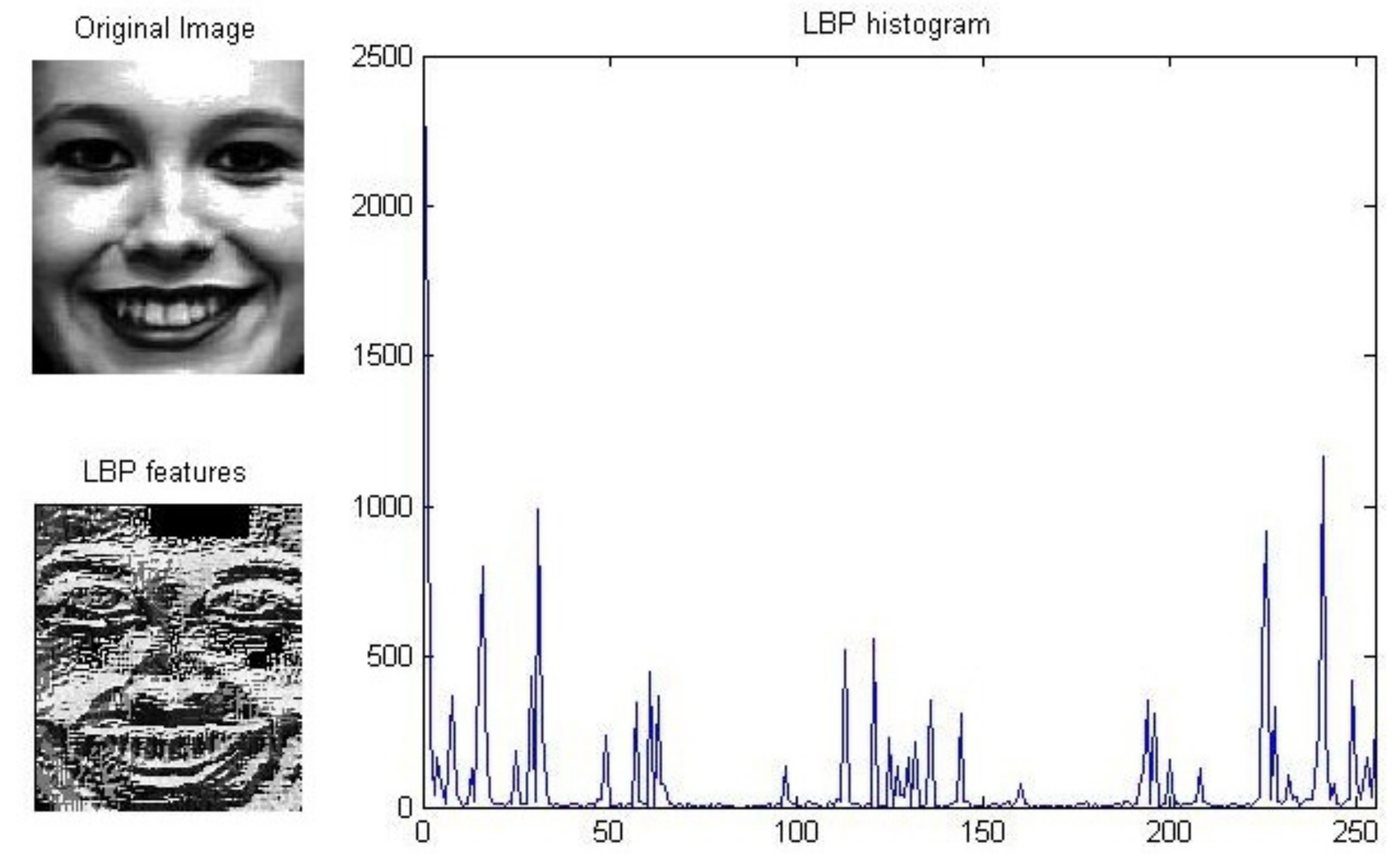}}
\subfigure[LBP histogram features for a face image with 5 blocks.] {\label{chap5-fig11}\includegraphics[scale=0.35]{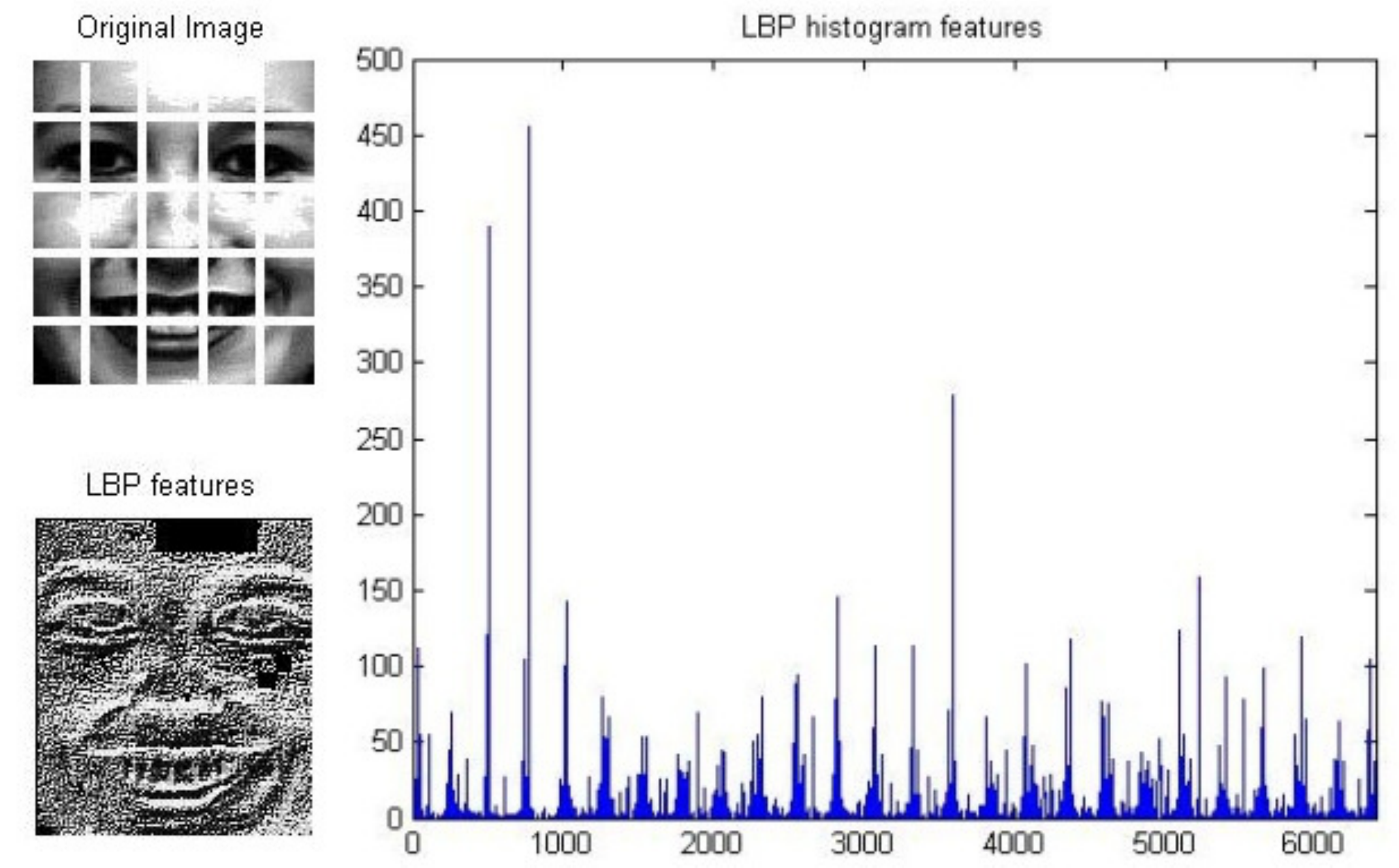}}
\end{center}
\caption{LBP histogram features of a face with and without blocks.}
\label{lbpfig}
\end{figure}

LBP feature extraction on a face image along with feature histogram is shown in figure \ref{chap5-fig10}. To increase the feature strength for more facial details, a face images are divided into number of blocks. Figure \ref{chap5-fig11} shows a face image with 5 number of division along row and column wise (totally 25 blocks) and its feature histogram. In this paper, each face image with 9 number of division along row and column wise (totally 81 blocks) are considered for experimental purpose. When a test image is given as input, the LBP histogram features are extracted which is then used for classification purpose. 

\subsection{Wavelet Decomposition (WD)}
\label{wd}
The wavelet transform offers the advantage of a sparse image representation and a complete representation \cite{wdref3}. Filter banks are elementary building blocks in the construction of wavelets. An Analysis filter bank consist of a low pass filter $H_0(e^{j\omega})$, a high pass filter $H_1(e^{j\omega})$ and down-samplers \cite{WDref1,WDref2,wdref4}. These filter banks are cascaded to form wavelet decomposition (WD). The decomposition can be performed on an image by first applying 1D filtering along rows of the images and then along columns, or vice versa. This is illustrated in the following figure \ref{wd1}.

\begin{figure}[htb]
\centering
\includegraphics[width=4in]{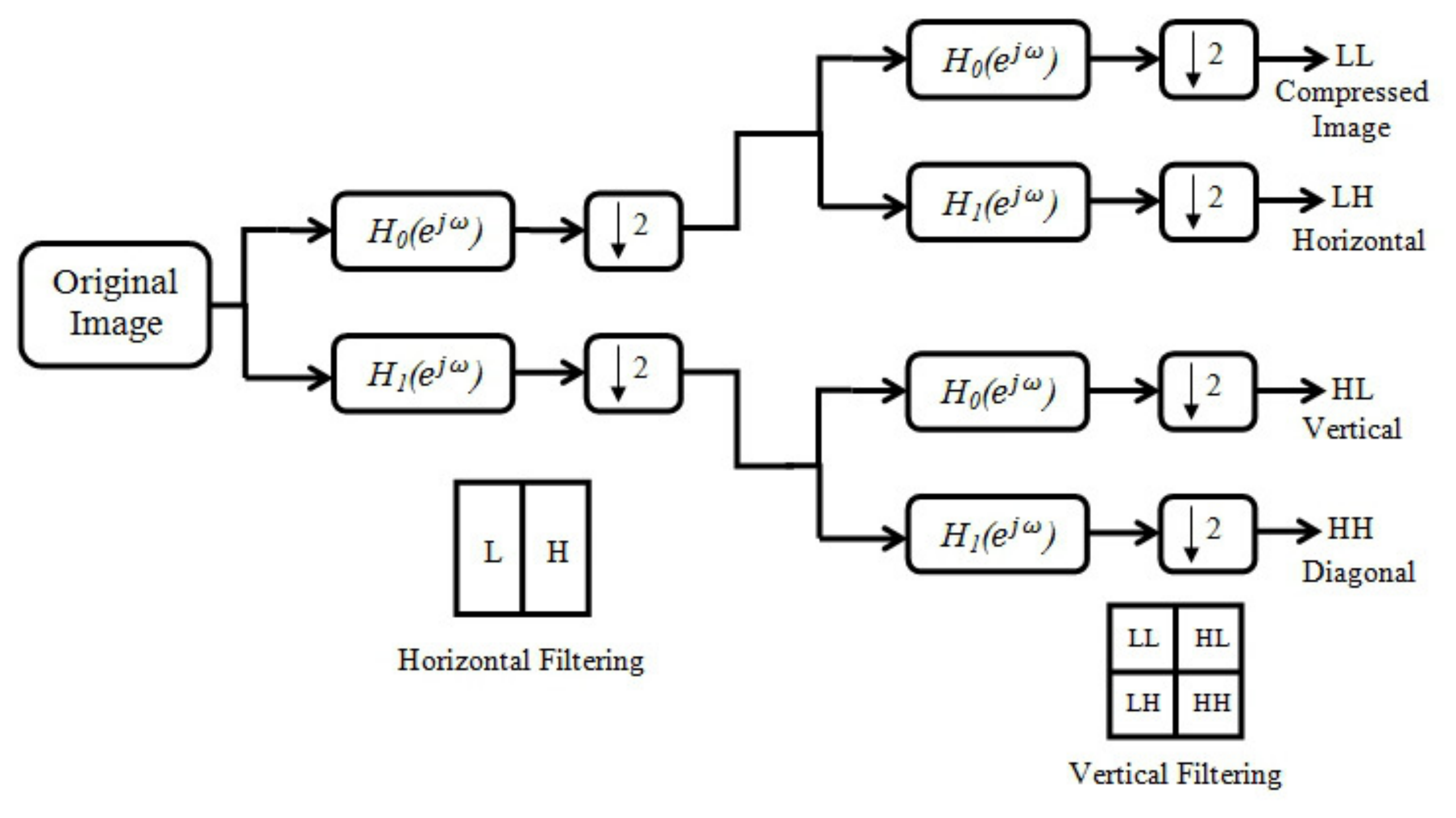}
\caption{The block diagram depicting the operation of wavelet decomposition.}
\label{wd1}
\end{figure}

\begin{figure}[htb]
\centering
\includegraphics[width=2.5in]{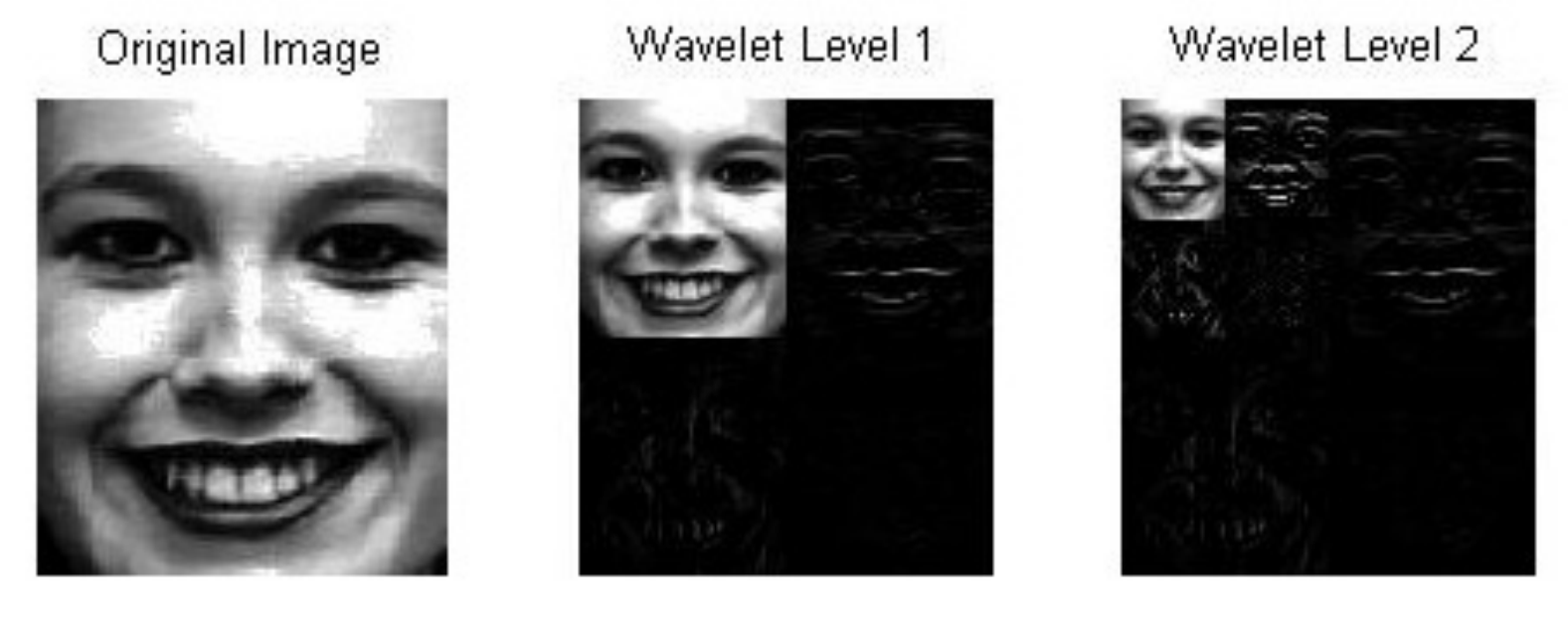}
\caption{Wavelet Decomposition of a face image.}
\label{wd2}
\end{figure}

The original image of size M x N pixels is processed along horizontal and vertical direction using low pass $H_0(e^{j\omega})$ and high pass filters $H_1(e^{j\omega})$. This 1D decomposition produces L and H which is also down-sampled to the rectangle of size M x N/2 pixels as shown in figure \ref{wd1}. These matrices are again transposed, decomposed and down-sampled along row wise to produce four subbands (or subimages) LL, HL, LH, HH of size M/2 x N/2 pixels. These subbands correspond to different spatial frequency bands in the image. The image with four subbands is called as wavelet level one. The LL component (i.e. compressed image) can be further decomposed to obtain LLLL, HLLL, LHLL, HHLL. This image with seven subbands is called as wavelet level two. This paper uses Daubechies wavelet 8 with two level of decomposition for all experiments. Figure \ref{wd2} shows a face image from Cohn-Kanade database \cite{cohndb} along with wavelet level 1 and level 2 decomposition. The wavelet level two image with seven subbands are concatenated to a 1D vector which is then passed to the feature dimension reduction step as shown in block diagram \ref{block2}.

\section{Feature Dimension Reduction using PCA}
\label{pca}
An image space is a space with number of pixels as dimensions. By converting image to an image vector using column concatenation, image can be treated as a point in the image space. When all the training set images are converted into vectors, they group at location with similar structure like eyes, nose and mouth with their relative position correlated. Eigenface method starts with this correlation and tries to find lower dimension space for the face images by focusing on the variation between face images.

The feature vectors (I) from Gabor/LBP/WD serves as the training set for the PCA method. Let M be the total number of images in the training set. The deviation of each image from the mean image is calculated using the equations (\ref{meanface}, \ref{meansub}). 
\begin{equation}
\label{meanface}
\psi= \frac{1}{M}\sum_{n=1}^MI_n
\end{equation}
\begin{equation}
\label{meansub}
\phi_n = I_n - \psi
\end{equation}
The variation among the training set images (i.e. eigenvectors) of the covariance matrix is calculated using equation (\ref{covariance}). The space where all this eigenvectors resides is called as eigenface space or eigenspace \cite{ref10}. All the training set images are projected into the eigenface space using equation (\ref{projection}). 
\begin{equation}
\label{covariance}
C = \frac{1}{M} \sum_{n=1}^M\phi_n\phi_n^T = AA^T
\end{equation}
\begin{equation}
\label{projection}
\omega_k = u_k . \phi = u_k . (I - \psi)
\end{equation}
Weight Matrix $\Omega = [\omega_1, \omega_2, \ldots , \omega_M']^T$
is the representation of a training image in the eigenface space. 

A new test image is classified by extracting gabor/LBP/WD features. It is then mean subtracted using equation (\ref{meansub}) followed by projection onto the eigenface space using equation (\ref{projection}). Weight matrix of the test image $\Omega_T = [\omega_1, \omega_2, \ldots , \omega_M']^T$ is calculated by projecting test image to eigenspace. This weight matrics $\Omega_T$ is used for classification purpose.

\section{Classification using Neural Networks}
\label{neural}
Neural Networks provides a great alternative to other conventional classifiers and decision making systems. In this paper, network consists of three layers and training is done using multilayer feedforward networks with gradient descent backpropagation algorithm. Number of input nodes is equal to the size of the feature vectors. Number of nodes in the hidden layer and number of iteration is experimental and it is discussed in section \ref{result}. 1's and -1's are used as target values for training appearance parameters (AAM features) whereas 1's and 0's for training other feature extraction methods. Tan-sigmoid is the transfer function used for both hidden as well as output layer. 0.0001 is set as goal for the network to achieve.

\section{Experimental Results and Discussions}
\label{result}
The results are separately discussed for gender classification, age estimation using gender information, expression recognition and racial recognition. This section also provides information about time taken for feature extraction, Neural training and testing an image.

\subsection{Gender Recognition}
\label{genrec}
The performance of the feature extraction methods for gender recognition is analyzed with Cohn-Kanade \cite{cohndb} and FG-NET \cite{fgnet}. From Cohn-Kanade database 880 images are considered out of which 463 images are used as training set and 417 images are used as test set. The training set consists of 239 male and 224 female face images. The test set consists of 208 male and 209 female face images. In order to check the gender classification rate in presents of aging effect, 321 images from FG-NET database are considered. Among 321 images, 218 images of which 111 male and 107 female face images are used as training set. Remaining 103 images of which 50 male and 53 female face images are used as test set. In AAM feature extraction, the texture size used is 200 x 200 pixels in case of Cohn-Kanade and 350 x 350 pixels for FG-NET database. To train the neural network, 500 numbers of hidden nodes and 5000 number of iterations are used for Cohn-Kanade database images. FG-NET face images are trained with 1000 hidden nodes and 6500 hidden layers. The results obtained from all four features for gender recognition is given in table \ref{table2}. Gabor feature performance is better than AAM for Cohn-Kanade database, but AAM performs well in the presents of aging variations i.e. in case of FG-NET database. The reason for increase in classification rate may be the shape landmark points as shown in figure \ref{fig3} and \ref{fig9a}. The shape landmark points provided along with FG-NET face database is more appropriate for gender classification and the one with Cohn-Kanade database is more suitable for expression recognition. The performance of LBP and Gabor are very similar.
\begin{table}[htp]
\begin{minipage}{0.43\textwidth}
\centering
\begin{tabular}{|c|c|c|}
\hline
\rowcolor[gray]{.8}
Methods & Cohn-Kanade & FG-NET\\
\hline
AAM & 90.795 & 92.523\\
\hline
Gabor & 91.30 & 90.03\\
\hline
LBP & 90.56 & 90.34\\
\hline
WD & 87.95 & 89.72\\
\hline
\end{tabular}
\captionof{table}{Results obtained from four feature extraction methods for gender recognition.}
\label{table2}
\end{minipage}
\hspace{0.3cm}
\begin{minipage}{0.52\textwidth}
\centering
\includegraphics[width=\textwidth]{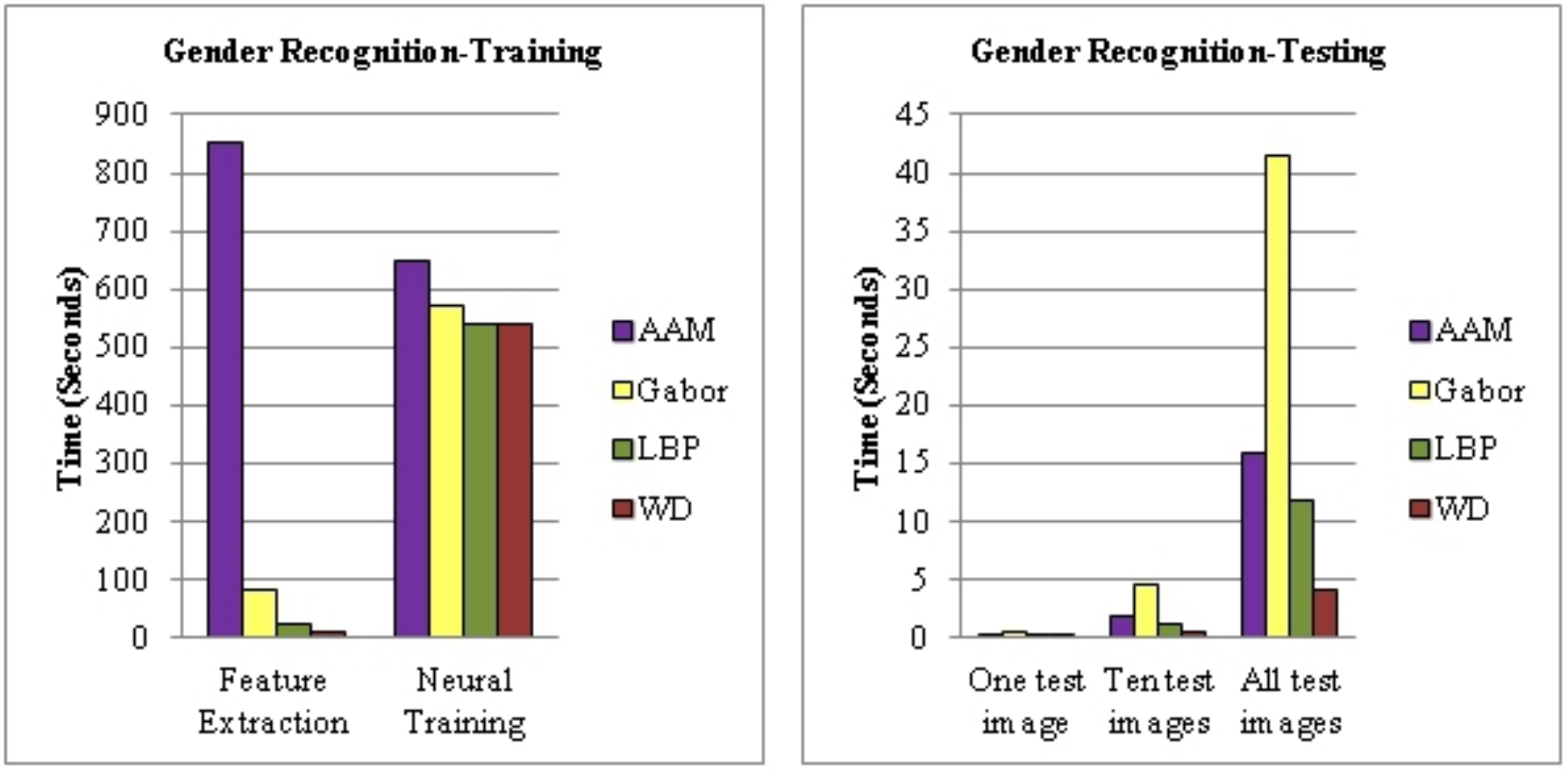}
\captionof{figure}{A comparison of time taken during training and testing four feature extraction methods.}
\label{table2fig}
\end{minipage}
\end{table}

The time taken for feature extraction and neural training for different methods are shown in figure \ref{table2fig}. It is clear that time taken for AAM feature extraction is higher than Gabor/LBP/WD and it applies for Neural training too. The time taken for testing one, ten, all (i.e. 103) images are also shown in figure \ref{table2fig}. This time include feature extraction and neural classification of the test set images. It is seen that time taken for computing AAM features are less compared to Gabor feature extraction. This is due to the time taken for convolving 40 Gabor filters. LBP and WD consume very less time compared to other two feature extractors. 

\subsection{Age Estimation using Gender Information}
\label{ageest}
Age estimation is analyzed with FG-NET \cite{fgnet} database. Totally 321 images are considered in which 218 images are used as training set and 103 images as test set. Age ranging from 0 - 60 is used for analysis and the number of male and female images are the same as in section \ref{genrec}. The texture size in case of AAM feature extraction is 350 x 350 pixels. The number of hidden nodes is 1000 for gender classification and 1200 for age estimation. The number of iteration is 6500 for gender classification and 8000 for age estimation. 
\begin{table}[htp]
\begin{minipage}{0.43\textwidth}
\centering
\begin{tabular}{|c|c|c|c|c|}
\hline
\rowcolor[gray]{.8}
Methods & AE & AEUGI\\
\hline
AAM & 72.27 & 77.25\\
\hline
Gabor & 70.40 & 75.70\\
\hline
LBP & 71.03 & 74.77\\
\hline
WD & 71.65 & 73.21\\
\hline
\end{tabular}
\captionof{table}{Results obtained from AE- Age Estimation and AEUGI- Age Estimation using Gender Information.}
\label{table3}
\end{minipage}
\hspace{0.3cm}
\begin{minipage}{0.52\textwidth}
\centering
\includegraphics[width=\textwidth]{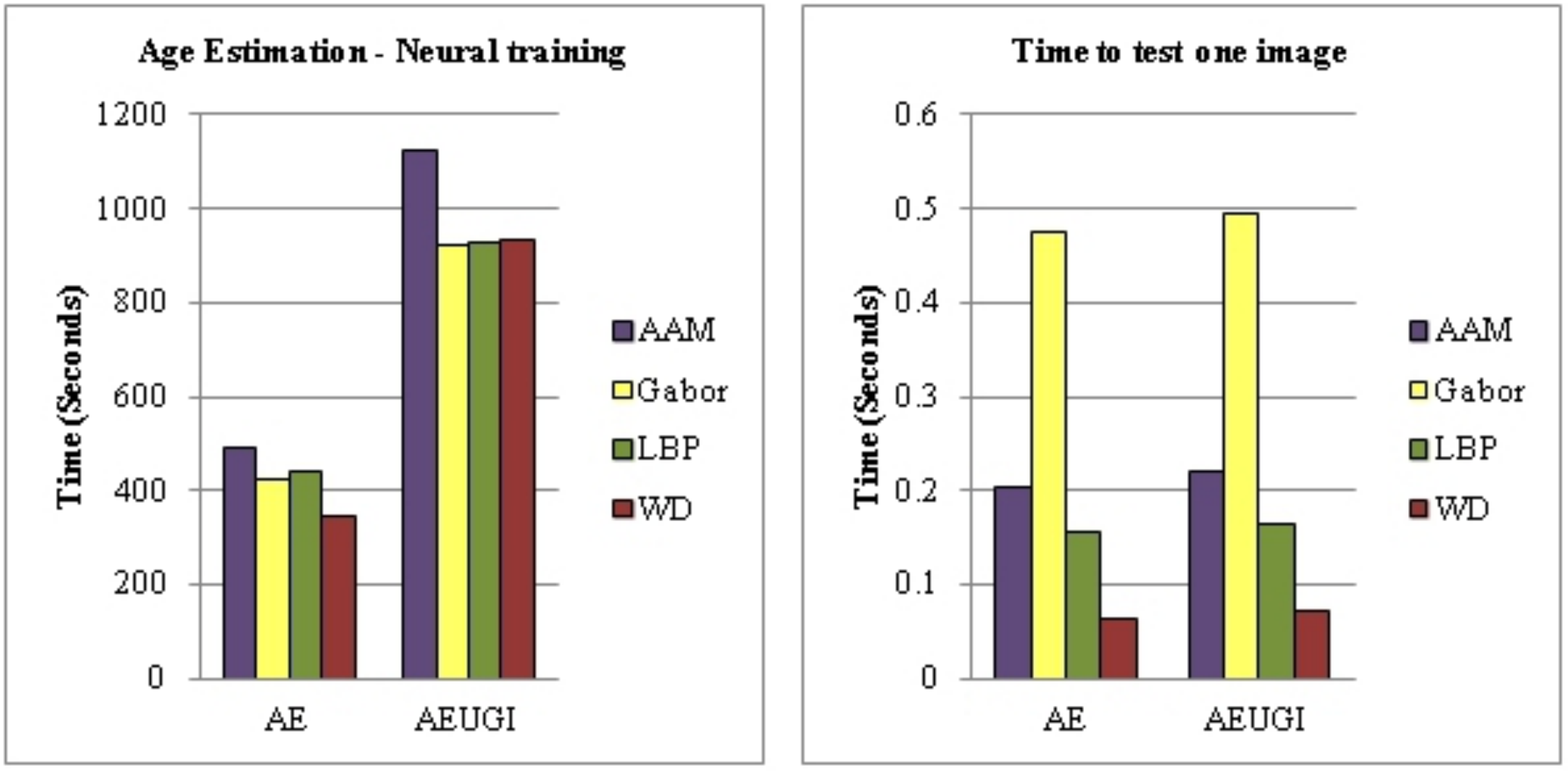}
\captionof{figure}{Time taken to train and test both AE and AEUGI using four features.}
\label{table3fig}
\end{minipage}
\end{table}

The results obtained using four different methods for age estimation and age estimation using Gender information is given in table \ref{table3}. Time taken to train neural network with all the four methods is shown in figure \ref{table3fig}. The performance of AAM is the best for both age estimation and age estimation using gender information than all other feature extraction methods. From chart, it is clear that time taken to train neural network for age estimation using gender information is higher than age estimation and in turn using AAM features takes more time than any other feature extractors. The time taken to test an image with age estimation using gender information and age estimation using all four features are also shown in figure \ref{table3fig}. Gabor features takes more time and WD takes least time to test an image than other methods for both AE and AEUGI. 

\subsection{Expression Recognition}
\label{exprerec}
\begin{figure}[htb]
\centering
\includegraphics[scale=0.45]{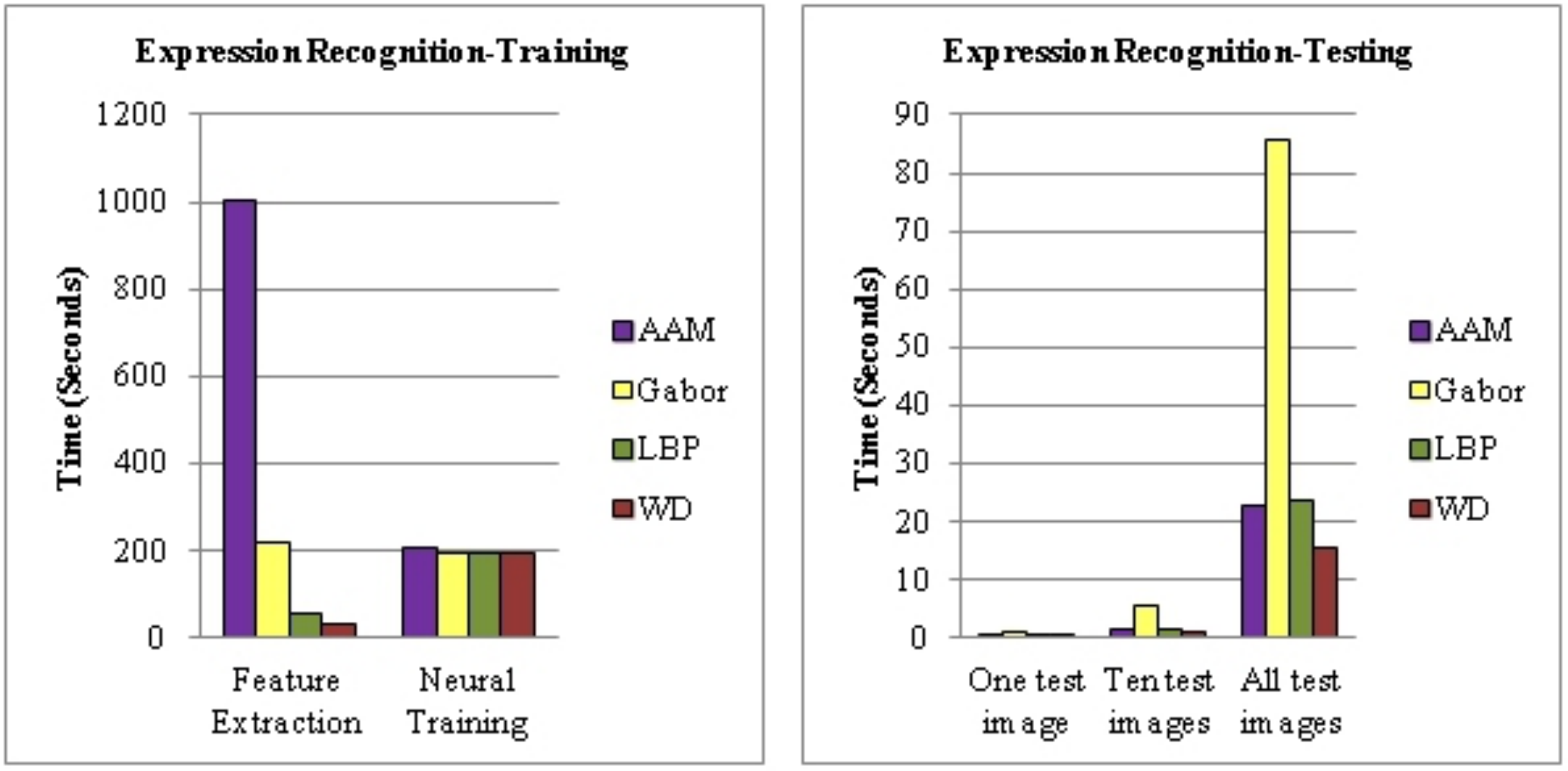}
\caption{Time taken for training and testing four different methods.}
\label{expresschart}
\end{figure}

Expression recognition is analyzed with Cohn-Kanade face database \cite{cohndb}. Totally 750 images with happy, angry, disgust, surprise, fear, sad expressions are considered. Among 750 images, 500 images are used as training set and 250 images are used as test set. Out of 500 training set images, fear and sad shares 50 images each and 100 images each from other four expression. In test set, fear and sad shares 25 images each and 50 images each from other four expression. The texture size used for AAM feature extraction is 150 x 150 pixels. In neural networks training, 200 hidden neurons with 5000 iterations are used. The results obtained using different features on Cohn-Kanade database is given in table \ref{table4}. The performance of AAM and LBP features gives better recognition rate compared to Gabor and WD features. The time taken for feature extraction, neural training and testing images is given in figure \ref{expresschart}. As mentioned in section \ref{genrec}, time taken for AAM feature extraction is more than all other methods and Gabor method takes more time in case of testing.

\begin{table}[htb]
\centering
\begin{tabular}{|c|c|c|c|c|c|c|c|}
\hline
\rowcolor[gray]{.8}
\backslashbox {Methods}{Expression} & Anger & Disgust & Fear & Happy & Sad & Surprise & Total\\
\hline
AAM & 94 & 94 & 90.6 & 96 & 94.67 & 100 & 95.33\\
\hline
Gabor & 84 & 86.67 & 57.33 & 94 & 72 & 100 & 85.88\\
\hline
LBP & 94 & 94 & 89.33 & 100 & 77.33 & 94 & 93.06\\
\hline
WD & 82 & 90 & 74.67 & 99.33 & 88 & 99.33 & 90.40\\
\hline
\end{tabular}
\caption{Results obtained from Expression recognition using Cohn-kanade database.}
\label{table4}
\end{table}

\subsection{Racial Recognition}
\label{race}
Acquiring ethnicity from a face image using all four methods are experimented with the face images from PAL \cite{paldb}, JAFFE \cite{jaffe} and FERET \cite{feretdb} databases. Four different ethnic groups namely white, black, indian and others (Asian, Hispanic) are considered. Totally 357 images from all three databases is used for training and testing purpose. Dataset consist of 80 images from white group, 90 from black group, 82 from indian group and 105 from others group. Among these images 40 from white, black, indian group and 50 from others group is used for training purpose (totally 170 images). Remaining 187 images are used for testing purpose. AAM feature extraction is performed with the texture size of 250 x 250 pixels. The neural network training is executed with 200 hidden layers and 5000 iterations. The results obtained using four different feature extraction methods for racial recognition is given in table \ref{tablerace}. AAM gives the best results for racial recognition. Time taken for training and testing is similar to the one explained in previous sections. 
\begin{table}[htp]
\begin{minipage}{0.43\textwidth}
\centering
\begin{tabular}{|c|c|c|c|c|}
\hline
\rowcolor[gray]{.8}
Methods & PAL+JAFFE+FERET\\
\hline
AAM & 93.83\\
\hline
Gabor & 90.19\\
\hline
LBP & 89.63\\
\hline
WD & 86.83\\
\hline
\end{tabular}
\captionof{table}{Results obtained from four different methods for racial recognition.}
\label{tablerace}
\end{minipage}
\hspace{0.3cm}
\begin{minipage}{0.52\textwidth}
\centering
\includegraphics[width=\textwidth]{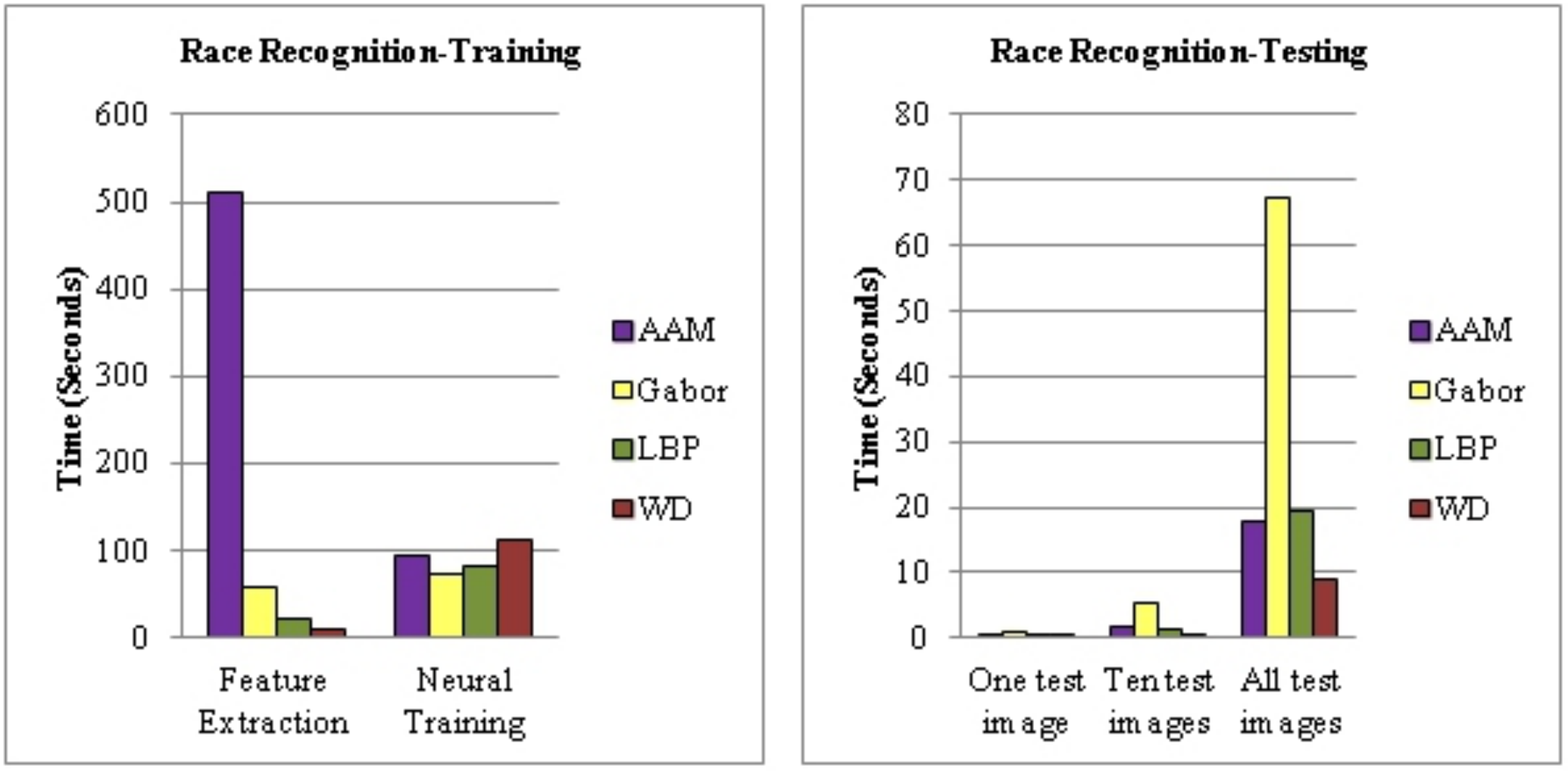}
\captionof{figure}{Time taken for training and testing using four features for racial recognition.}
\label{figrace}
\end{minipage}
\end{table}

The AAM features for gender recognition, age estimation, expression recognition and racial recognition gives good result compared to other three features. The performance of Gabor and LBP are very similar in all cases whereas LBP consumes very less time for feature extraction and testing than Gabor. Wavelet decomposition (WD) provides least recognition rate in most of the experiments, but time to extract features and test an image is very less than all other methods. The time taken for feature extraction and neural training is higher in case of AAM features than others. In most of the situations training takes place in offline, so even longer training duration is agreeable. The time to test the image is all matters, Gabor takes more time to test an image than other methods. This is due to the complexity in convolving 40 gabor filters. Comparing accuracy and time to test an image AAM is better suitable for real time application provided with the shape landmark points. There are methods to iteratively find the shape landmark points which is not analyzed in this paper. Some applications may need very less training as well as testing time, even less accuracy can be acceptable. This condition is better provided by wavelet decomposition. Particular method will be best suitable for particular application and environment, hence it is very hard to mention a feature extractor which is best in all circumstances.

\subsection{Gender, Age, Expression, Ethnicity from an image}
\label{comb}
\begin{table}[htp]
\begin{center}
\begin{minipage}{0.28\textwidth}
\includegraphics[width=\textwidth]{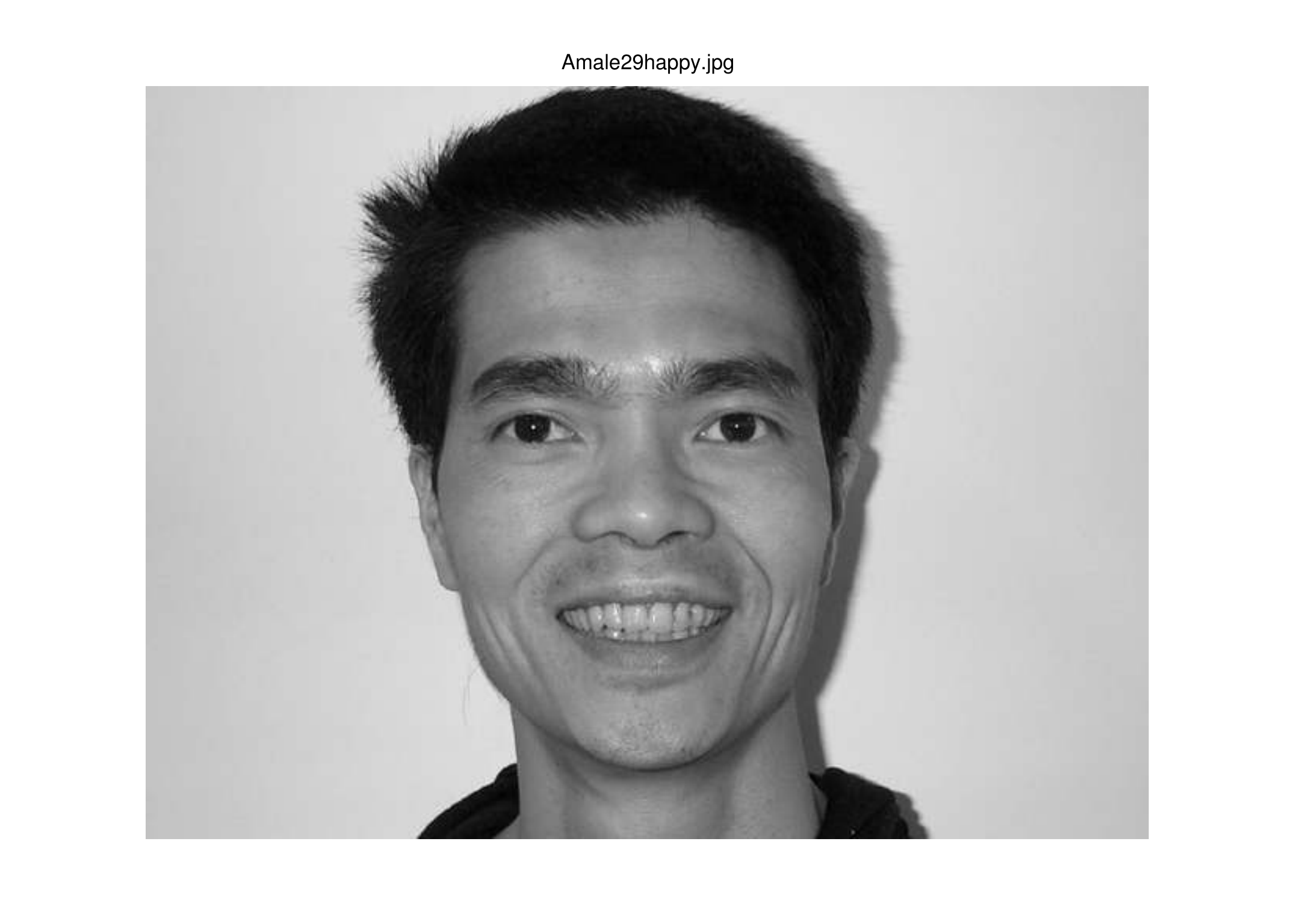}
\end{minipage}
\hspace{0.5cm}
\begin{minipage}{0.5\textwidth}
\begin{tabular}{|c|c|c|c|c|}
\hline
\rowcolor[gray]{.8}
Actual & AAM & Gabor & LBP & WD\\
\hline
Male & Male & Male & Male & Male\\
\hline
29 & 20-30 & 20-30 & 10-20 & 10-20\\
\hline
Happy & Happy & Surprise & Happy & Happy\\
\hline
Other & Other & Other & other & white\\
\hline
\end{tabular}
\end{minipage}
\begin{minipage}{0.28\textwidth}
\includegraphics[width=\textwidth]{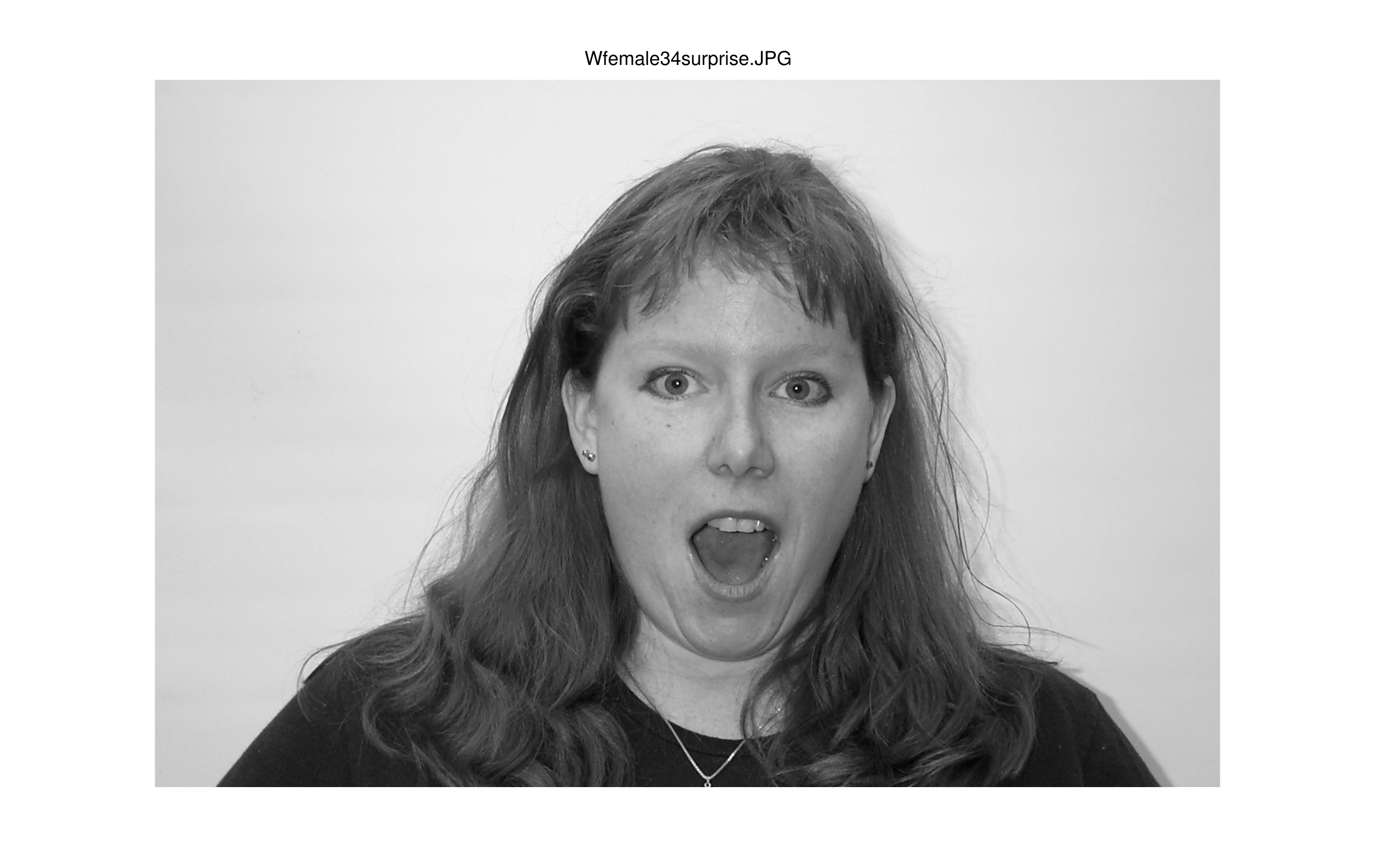}
\end{minipage}
\hspace{0.5cm}
\begin{minipage}{0.5\textwidth}
\begin{tabular}{|c|c|c|c|c|}
\hline
\rowcolor[gray]{.8}
Actual & AAM & Gabor & LBP & WD\\
\hline
Female & Female & Female & Female & Male\\
\hline
34 & 30-40 & 20-30 & 30-40 & 40-50\\
\hline
Surprise & Surprise & Happy & Suprise & Happy\\
\hline
white & white & Others & White & Indian\\
\hline
\end{tabular}
\end{minipage}
\begin{minipage}{0.28\textwidth}
\includegraphics[width=\textwidth]{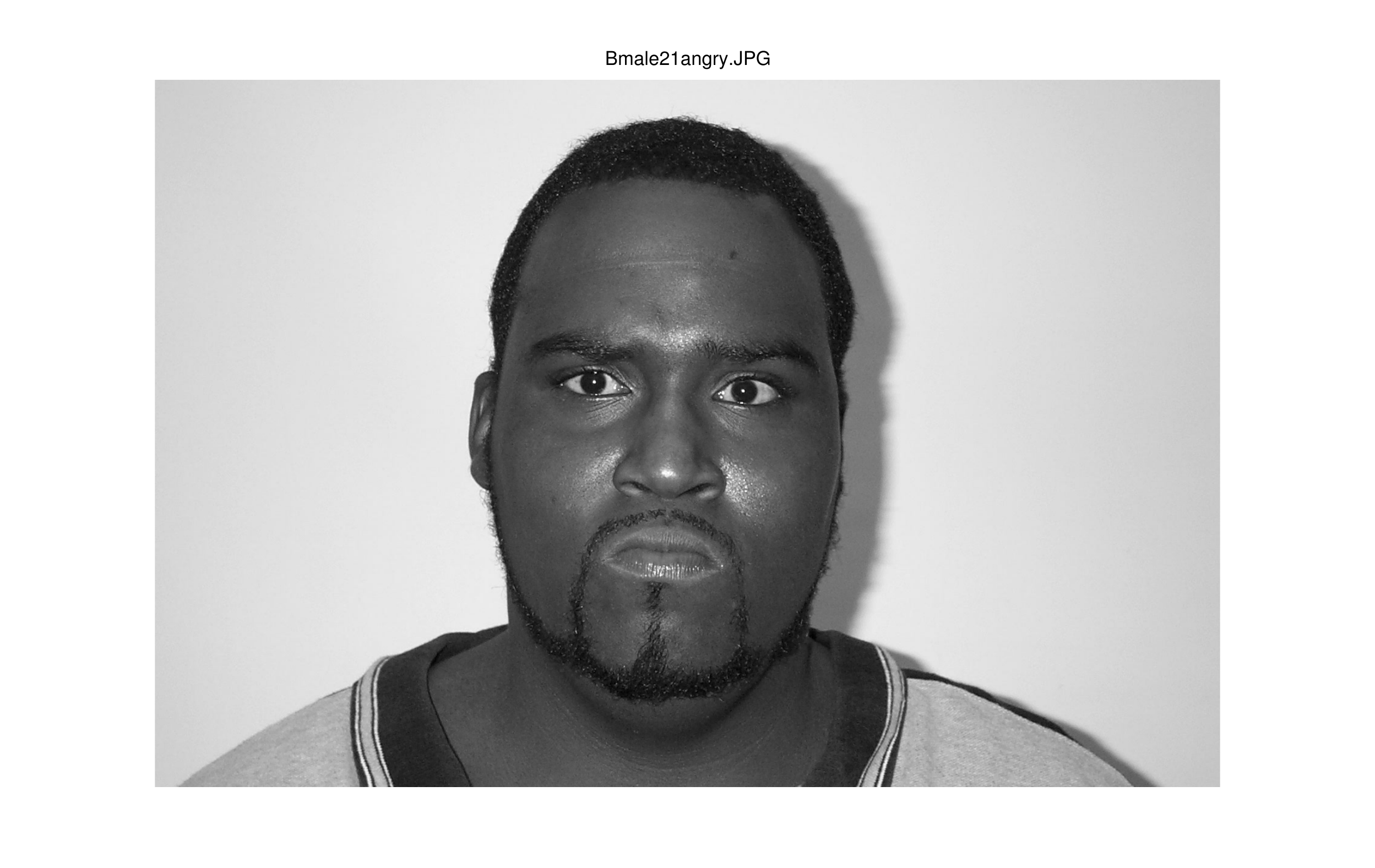}
\end{minipage}
\hspace{0.5cm}
\begin{minipage}{0.5\textwidth}
\begin{tabular}{|c|c|c|c|c|}
\hline
\rowcolor[gray]{.8}
Actual & AAM & Gabor & LBP & WD\\
\hline
Male & Male & Male & Male & Male\\
\hline
21 & 10-20 & 20-30 & 10-20 & 30-40\\
\hline
Angry & Angry & Angry & Angry & Angry\\
\hline
Black & Black & Black & Black & Black\\ 
\hline
\end{tabular}
\end{minipage}
\begin{minipage}{0.28\textwidth}
\includegraphics[width=\textwidth]{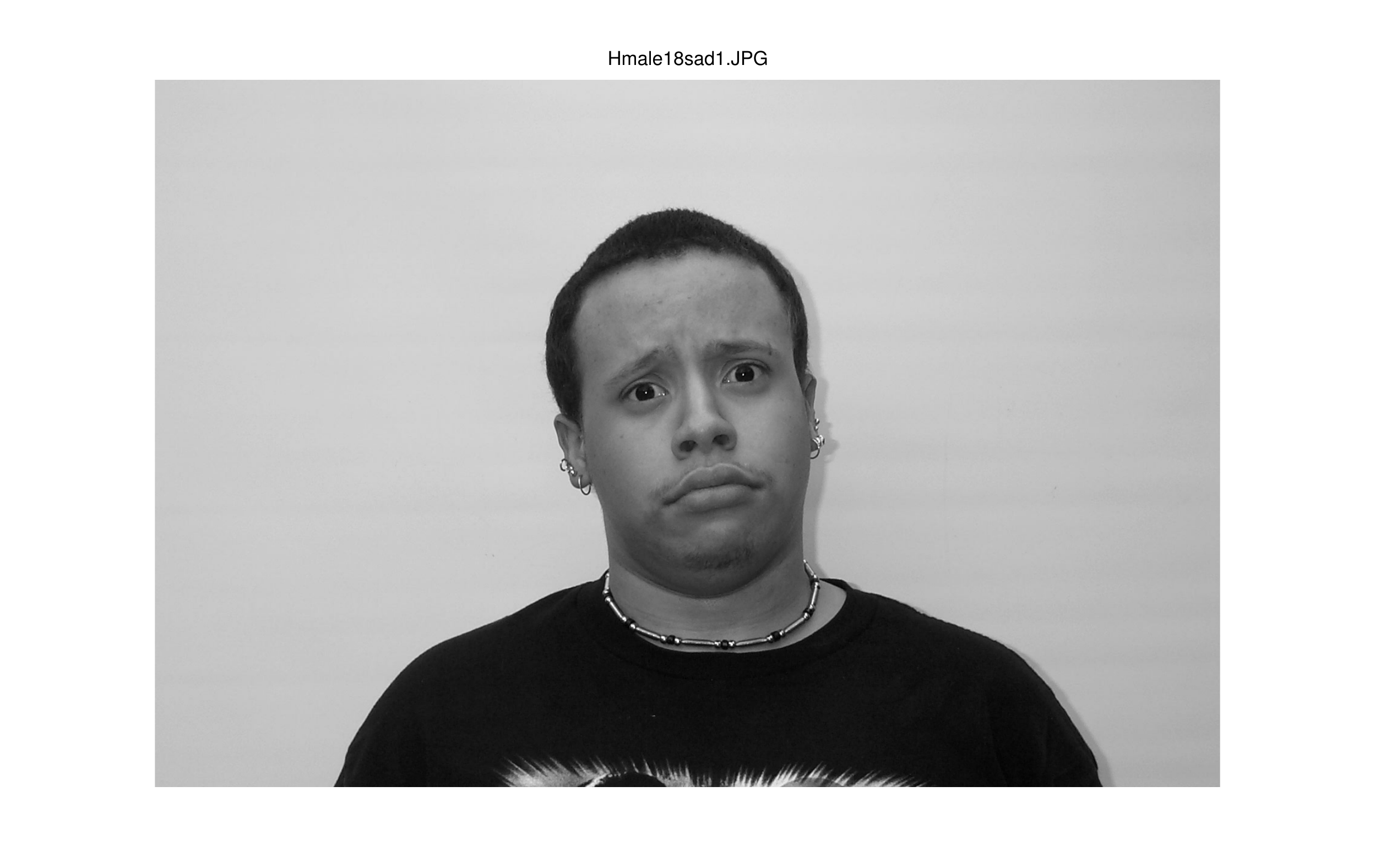}
\captionof{figure}{Test image.}
\label{refig2}
\end{minipage}
\hspace{0.5cm}
\begin{minipage}{0.5\textwidth}
\begin{tabular}{|c|c|c|c|c|}
\hline
\rowcolor[gray]{.8}
Actual & AAM & Gabor & LBP & WD\\
\hline
Male & Male & Male & Male & Female\\
\hline
18 & 10-20 & 0-10 & 10-20 & 0-10\\
\hline
Sad & Fear & Fear & Sad & Happy\\
\hline
Other & Other & Other & Other & Other\\
\hline
\end{tabular}
\captionof{table}{Actual and Analyzed data.}
\label{retable2}
\end{minipage}
\end{center}
\end{table}
It is essential to obtain gender, age, expression (or mood) and ethnicity (or racial) information from a face image to automate applications as well to bring in all the human abilities in a machine. This section concentrates on obtaining all these information from a face image. Experiments are performed by combining Gender recognition, Age estimation, expression recognition and racial recognition using all the above mentioned four different feature extractors. This combination is executed by using trained neural network from gender recognition in section \ref{genrec}, Age estimation using Gender Information in section \ref{ageest}, expression recognition in section \ref{exprerec} and racial recognition in section \ref{race}. In real time condition, the images used for training and testing will be taken under different environment. This is also analyzed by performing training with different database images as detailed in previous sections and testing is performed with the images of PAL database \cite{paldb}. The results obtained are shown along with the test image and analyzed data in figure \ref{refig2} and table \ref{retable2} respectively.

The time taken for the first image is less than a second with AAM, LBP and WD methods whereas Gabor taken more than a second to complete the same task. Time shown here in all cases is acquired with 4 GB RAM and 2.40 GHz speed processor using MATLAB 7.0 software. In third case, the actual age is 21 but the analyzed results for age are between 10 to 20 years. This can be overcome by providing more training images and increasing the number of age divisions into many ranges. In fact age prediction is difficult even with human begins, since each person has different way of growth at different stage. In last case, the expression is recognized incorrectly, this is due to the expression defined by different database images looks totally different. As well different ethnic group may express their emotions in different way. This is the major problem to be addressed in the real time situation. Still the efficiency can be improved by considering images from different expression recognition databases. 

\section{Conclusion}
\label{conclude}
A detailed comparison of AAM, Gabor, LBP and WD features for gender recognition, age estimation using gender information, expression recognition and racial recognition is provided along with recognition rate, time taken for feature extraction, neural training and for testing an image. Results shows that AAM features are better than other features in terms of accuracy and time taken for testing an image. LBP and Gabor gives similar performance, whereas LBP is computationally less expensive. In term of time consumption during training and testing, WD is better than other methods. Aging effect in case of gender recognition can be tackled using AAM features. The performance of gender recognition is affected by using different shape landmark points which shows the inconsistence of AAM features. The accuracy of age estimation is improved by cascading gender information. Finally an attempt has been made in combining gender recognition, age estimation, expression recognition and racial recognition. Retrieving all (gender, age range, expression and ethnicity) information from a face image in less than a second time duration using AAM, LBP and WD features is also demonstrated. 



\bibliographystyle{model1b-num-names}
\bibliography{patternrecognitionpaper}







\end{document}